\documentclass[letterpaper, 10 pt, conference]{ieeeconf}  % Comment this line out if you need a4paper

\IEEEoverridecommandlockouts                              % This command is only needed if 
\usepackage{amsmath} % assumes amsmath package installed
\usepackage[nolist,nohyperlinks]{acronym}
\usepackage{todonotes}
\usepackage{graphicx}
\usepackage{subcaption}
\usepackage{amsfonts} 
\usepackage{float}
\usepackage{comment}
\usepackage{siunitx}
\usepackage[ruled]{algorithm2e}
\SetKwComment{Comment}{/* }{ */}

\acrodef{AI}{Artificial Intelligence}
\acrodef{CAD}{Computer-Aided Design}
\acrodef{NeRF}{Neural Radiance Field}
\acrodef{NGP}{Neural Graphics Primitives}
\acrodef{DoF}{Degrees of Freedom}
\acrodef{SDF}{Signed Distance Field}
\acrodef{MLP}{Multi-Layer Perceptron}

\newcommand{\point}{\mathbf{x}}

\newcommand{\Rthree}{\mathbb{R}^3}
\newcommand{\R}{\mathbb{R}}

\newcommand{\SOthree}{\mathbb{SO}(3)}

\newcommand{\pose}{\xi}

\newcommand{\densityf}{\sigma}

\newcommand{\normal}{\mathbf{n}}
\newcommand{\normalset}{\mathcal{N}}

\newcommand{\surface}{\mathcal{S}}

\newcommand{\sample}{s}

\newcommand{\hypothesesset}{\mathcal{H}}
\newcommand{\hypothesis}{\xi}

\newcommand{\model}{\mathcal{M}}
\newcommand{\pointset}{\mathcal{X}}
\newcommand{\pointsurfaceset}{\pointset^\surface}
\newcommand{\pointnormalset}{\pointset^\normalset}
\newcommand{\pointsurface}{\point^\surface}
\newcommand{\pointnormal}{\point^\normalset}
\newcommand{\fitnessfunc}{f}

\newcommand{\reffig}[1]{Fig.~\ref{#1}}
\newcommand{\refsec}[1]{Sec.~\ref{#1}}
\newcommand{\reftab}[1]{Tab.~\ref{#1}}
\newcommand{\refeq}[1]{Eq.~(\ref{#1})}

\newcommand{\ignore}[1]{}

\newcommand{\ba}{\begin{array}}
\newcommand{\ea}{\end{array}}
\newcommand{\bc}{\begin{center}}
\newcommand{\ec}{\end{center}}
\newcommand{\be}{\begin{enumerate}}
\newcommand{\ee}{\end{enumerate}}
\newcommand{\bea}{\begin{eqnarray}}
\newcommand{\eea}{\end{eqnarray}}
\newcommand{\beas}{\begin{eqnarray*}}
\newcommand{\eeas}{\end{eqnarray*}}
\newcommand{\beq}{\begin{equation}}
\newcommand{\eeq}{\end{equation}}
\newcommand{\bfig}{\begin{figure}}
\newcommand{\efig}{\end{figure}}
\newcommand{\bi}{\begin{itemize}}
\newcommand{\ei}{\end{itemize}}
\newcommand{\bpic}{\begin{picture}}
\newcommand{\epic}{\end{picture}}
\newcommand{\btabular}{\begin{tabular}}
\newcommand{\etabular}{\end{tabular}}
\newcommand{\btable}{\begin{table}}
\newcommand{\etable}{\end{table}}

\newcommand{\es}{\vfill
                 \rule[-6mm]{170mm}{0.7mm} \\
                 \redw{{\tiny
		  \hfill S-\theslide}}
                 \end{slide}}

\providecommand{\etal}{{\em et~al.}}

\def \hbar {{\bar{h}}}

\makeatletter
\renewcommand*\env@matrix[1][*\c@MaxMatrixCols c]{%
  \hskip -\arraycolsep
  \let\@ifnextchar\new@ifnextchar
  \array{#1}}
\makeatother

\title{\LARGE \bf
Fit-NGP:
Fitting Object Models to Neural Graphics Primitives 
}

\author{Marwan Taher, Ignacio Alzugaray and Andrew J. Davison\\
Dyson Robotics Lab, Imperial College London\\
{\tt\small \{m.taher, i.alzugaray, a.davison\}@imperial.ac.uk}% <-this % stops a space
}

\begin{document}

\maketitle
\thispagestyle{empty}
\pagestyle{empty}

%%%%%%%%%%%%%%%%%%%%%%%%%%%%%%%%%%%%%%%%%%%%%%%%%%%%%%%%%%%%%%%%%%%%%%%%%%%%%%%%
\begin{abstract}
Accurate 3D object pose estimation is key to enabling many robotic applications that involve challenging object interactions. In this work, we show that the density field created by a state-of-the-art efficient radiance field reconstruction method 
is suitable for highly accurate and robust pose estimation for objects with known 3D models, even when they are very small and with challenging reflective surfaces. We present a fully automatic object pose estimation system based on a robot arm with a single wrist-mounted camera, which can scan a scene from scratch, detect and estimate the 6-\ac{DoF} poses of multiple objects within a couple of minutes of operation. Small objects such as bolts and nuts are estimated with accuracy on order of 1mm.
\end{abstract}

%%%%%%%%%%%%%%%%%%%%%%%%%%%%%%%%%%%%%%%%%%%%%%%%%%%%%%%%%%%%%%%%%%%%%%%%%%%%%%%%
\section{Introduction}

% Andy editing: original text by Ignacio available in offcuts.tex

It remains a significant challenge to enable robots to manipulate objects around them with enough competence to unlock applications such as general domestic robotics, especially when these robots must rely only on their own on-board sensors such as cameras.
While simple picking up and dropping can often be achieved via direct image-to-action control policies, more complex manipulation such as precise placing or insertion, can benefit from explicit reasoning about the 3D shape of objects.

While general object shape estimation is an interesting and important problem, in most application scenarios (e.g. office, factory, kitchen or household) a robot will usually be dealing with objects whose type is known in advance. 
Precise 3D models are often also available 
in the form of \ac{CAD} provided by the manufacturer, shared by other robots or estimated from the robot's own past experiences. 
In this case, 3D scene understanding takes the form of {\em pose estimation} of known object models.

In this paper, we show for the first time that the recent real-time light field reconstruction method in the form of Instant-NGP \cite{instant_ngp} is ready to be straightforwardly used as an intermediate representation as part of a fully automatic system for highly accurate 3D object pose estimation in table-top settings for precise manipulation tasks.
Our system comprises a robot arm with a single wrist-mounted RGB camera, which makes a rapid scan, reconstructs the scene, and fits object models all within two minutes of operation.

\begin{figure}[t]
    \centering
    \includegraphics[width=0.48\textwidth]{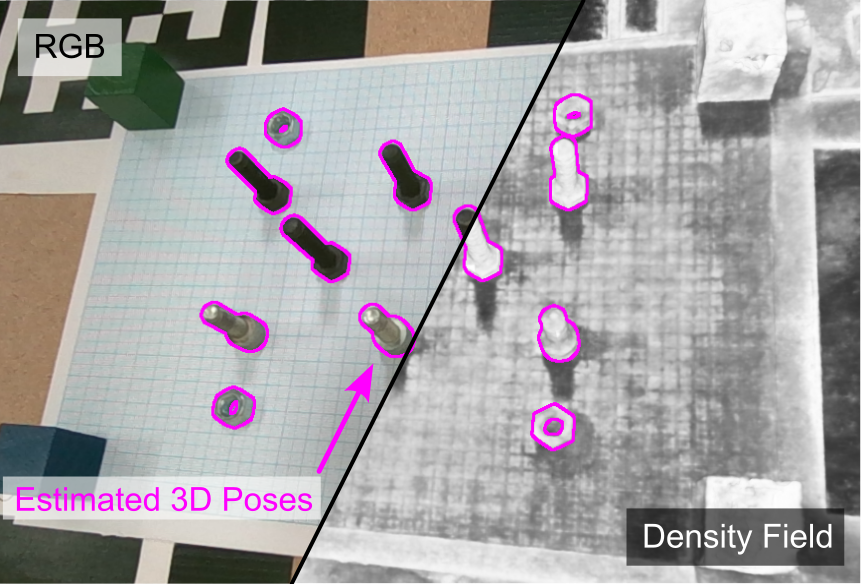}
    \caption{A set of posed images (top-left) from a scene containing multiple objects is used to train a \ac{NeRF}. 
    The reconstructed density field (bottom-right) is employed to align model-to-scene poses for each object using multi-hypothesis optimisation, with the best pose silhouette reprojection overlaid (magenta). }
    \label{fig:introduction}
    \vspace{-0.5cm}
\end{figure}

RGB light field reconstruction was recently revitalised by \ac{NeRF} \cite{Mildenhall:etal:ECCV2020}, which uses a single neural network optimised via volume rendering to reconstruct scene density and illumination. 
Instant-NGP (Instant Neural Graphics Primitives) \cite{instant_ngp} is a development which uses a much more efficient hybrid grid/neural representation than \ac{NeRF} to achieve efficient optimisation and rendering. 
Since it was designed primarily for visual fidelity rather than geometry estimation, 
the reconstructed scene density field is not necessarily accurate nor smooth. However, here we show that it contains details which are sufficiently suitable for object pose estimation, especially when heavily relying on object edges. 
We perform pose estimation via straightforward iterative optimisation of a cost function measuring the agreement between an object model and the Instant-NGP density field (see \reffig{fig:introduction}). 
Object poses are initialised automatically using off-the-shelf RGB object detection. 
Our full system includes automatic scene scanning, with camera pose estimates coming initially from robot arm kinematics and then refined using Instant-NGP's camera pose optimisation function for increased accuracy.
The use of kinematics means that the reconstruction is correctly scaled and object poses can be estimated with high metric precision.

This approach has many advantageous properties for object pose estimation in a robot manipulation setting. Firstly, it requires only a single RGB camera which the robot moves by itself. 
Alternatives such as depth cameras or stereo rigs typically have minimum and maximum range limits and limited reconstruction precision, as well as usually being more bulky and expensive. 
A single RGB camera can be moved arbitrarily close to the scene to gather great scene detail. 
We will show that this means that the pose of even tiny objects such as bolts and nuts can be estimated with an accuracy of less than 1.5 mm.
Further, light field estimation methods can cope with a wide range of lighting conditions and can build density maps which enable pose estimation even of objects which are usually difficult to deal with in computer vision due to issues such as shiny or metallic surfaces. Our method requires only geometric shape models of the objects, without any colour or texture information.

Fit-NGP is simple, accurate, and automatic, and is a method that other researchers will easily be able to copy and use in a wide range of manipulation settings.
We present results which demonstrate the accuracy of the method in real robot experiments, and its ability to deal with different objects and lighting conditions.

%In summary, the contributions of this paper are as follows:
%\begin{itemize}
%    \item TO BE UPDATED ON NEW EXPERIMENTS / FOCUS.
%    \item A 6DOF pose estimation method leverages Nerf as the under-laying scene reconstruction and accurately fits known models to it.
%    \item Real-world experiments on a robot arm demonstrating the framework capabilities, such as dealing with different lighting conditions, occlusions and specular/reflective objects.
%    \item Experiments on pose estimation datasets, showcasing state of the art results on the accuracy of pose estimation.
%\end{itemize}

%%%%%%%%%%%%%%%%%%%%%%%%%%%%%%%%%%%%%%%%%%%%%%%%%%%%%%%%%%%%
\section{Related Work}
6-\ac{DoF} pose estimation of objects is a long-standing problem in computer vision, and much work has focused on methods which only require one input image, from classical pipelines using RANSAC+PnP \cite{pnp} up to recent learning-based methods such as PoseCNN \cite{posecnn},  DeepIM \cite{deepim}, or MegaPose \cite{megapose}.

Single-view methods are fundamentally sensitive to occlusions, poor lighting conditions or ambiguities.
In manipulation-oriented setups where a robot controls the motion of a camera on its end effector, a robot can rapidly capture multiple frames while moving, and use all of the information to aid pose estimation. 
One option is  to pool many single-frame pose estimates via multi-view constraints as in
CosyPose \cite{cosypose}. 
Similar multi-view refinement has been taken to the scale of whole scenes in the object-based SLAM literature characterised by works such as SLAM++ \cite{Salas-Moreno:etal:CVPR2013}.%, Fusion++  \cite{McCormac:etal:3DV2018} or NodeSLAM \cite{Sucar:etal:3DV2020}.
% Fusion\texttt{++} \cite{McCormac:etal:3DV2018} represented the scene using online TSDF reconstructions of objects in the scene and NodeSLAM \cite{Sucar:etal:3DV2020} used optimisable multi-class learned object descriptors obtaining object reconstructions.

Alternatively, multiple frames can be used to
build an intermediate 3D scene representation before attempting pose estimation directly against the reconstruction, which is the approach we follow in this paper.
Related approaches designed for a manipulation setting include
MoreFusion \cite{Wada:etal:CVPR2020} which used a wrist-mounted depth camera to perform octree-based occupancy reconstruction \cite{octomap} before fitting object CAD models and performing
collision-aware pose refinement to deal with piles of objects. 
Scan2CAD
\cite{Avetisyan:etal:CVPR2019} achieved something similar at a room scale using 3D CNNs for alignment.

During manipulation, cameras are close to the scene and the reconstruction accuracy and minimum range of depth cameras can often become a problem.
New developments in light field estimation offer new possibilities to use RGB cameras to build a more accurate intermediate representation. 
\ac{NeRF} \cite{Mildenhall:etal:ECCV2020} 
made a breakthrough by showing that a coordinate-based \ac{MLP} can be trained through volume rendering to produce a photo-realistic scene representation from a set of posed RGB views and with no need for prior information, though at the cost of expensive off-line processing.
iMAP \cite{Sucar:etal:ICCV2021} was the first real-time capable scene modelling system based on a NeRF-like MLP, but the requirement for a depth camera and a cut-down network size for speed meant that its reconstruction accuracy was not suitable for object pose estimation.

Recently, hybrid representations have emerged that train much faster than NeRF, and achieve higher view synthesis quality, 
especially Instant-NGP, which uses multi-resolution hash encoding of 3D voxel grids, indexing small \acp{MLP}, and converges in tens of seconds for many scenes.  

\ac{NeRF} and Instant-NGP were designed for high fidelity view synthesis, not accurate reconstruction, and the density fields they reconstruct are often noisy.
It is possible to apply a regularisation prior
\cite{Oechsle:etal:ICCV2021} to improve surface smoothness. 
Instead, in our work we directly apply the strongest prior available --- that the world is made of up objects for which we have known models --- and directly fit these models against the raw density reconstruction. 
Although fuzzy in places, we have found that the reconstructions include accurate details on edges and high texture regions which allow extremely accurate object alignment, even for small objects with reflective surfaces which are very difficult to deal with in most RGB view-based methods. 
Instant-NGP can cope with these challenging issues and allows pose estimation to be purely based on the models and scene geometry.

NeRF is already beginning to be used in some robot systems such as in Dex-\ac{NeRF} \cite{dex_nerf} and Evo-\ac{NeRF} \cite{evp_nerf}.
There is some work on aligning multiple \ac{NeRF} reconstructions such as nerf2nerf \cite{nerf2nerf}, but we are not aware of other work attempting the alignment of object models against them.

\begin{comment}
ROCA: Robust CAD Model Retrieval and Alignment from a Single Image?

- Multi-view

* Vid2CAD: CAD Model Alignment using Multi-View Constraints from Videos?

6D Pose Estimation for Textureless Objects on RGB Frames using Multi-View Optimization; They decouple the translation and rotation estaimtion.

MVTrans: Multi-View Perception of Transparent Objects
Multi-View Object Pose Refinement with Differentiable Renderer

- Intermediate reconstruction for pose estimation

ReorientBot \cite{Wada:etal:ICRA2022b}

- Different types of intermediate reconstruction

- Neural radiance fields

- Regularising Nerf for reconstruction?

- Registration against Nerf?

DReg-Nerf: Deep Registration for Neural Radiance Fields; Nerf2Nerf flavoured registration but without human annotation. Use a pre-trained transformer to align so not relevant???

* Zero Nerf: Registration with Zero Overlap
* BundleSDF: Neural 6-DoF Tracking and 3D Reconstruction of Unknown Objects; Mention them as method that uses neural field for pose estimation/tracking?

- Nerf in robotics:
* Dex-Nerf: Using a Neural Radiance Field to Grasp Transparent Objects
 *Evo-Nerf: Evolving Nerf for Sequential Robot Grasping of Transparent Objects

% Andy: not sure these are relevant

%Indirect Point Cloud Registration: Aligning Distance Fields Using a Pseudo Third Point Set normals
%SDFReg: Learning Signed Distance Functions for Point Cloud Registration
% RGB-Only Reconstruction of Tabletop Scenes for Collision-Free Manipulator Control
\end{comment}

%%%%%%%%%%%%%%%%%%%%%%%%%%%%%%%%%%%%%%%%%%%%%%%%%%%%%%%%%%%%
\section{Methodology}

At its most general, Fit-NGP enables pose retrieval of multiple objects in an arbitrary scene given a set of RGB images with approximate camera poses, 3D models of the shape of the objects, and an off-the-shelf image segmentor capable of identifying the objects in one of these images, requiring no additional training data.
We demonstrate a system designed for a potential indoor manipulation, where images are captured by an automatic scan from a single wrist-mounted RGB camera on a robot arm, with approximate camera poses coming from the arm's known kinematics.

\begin{figure}[t]
    \centering
    \includegraphics[width=0.48\textwidth]{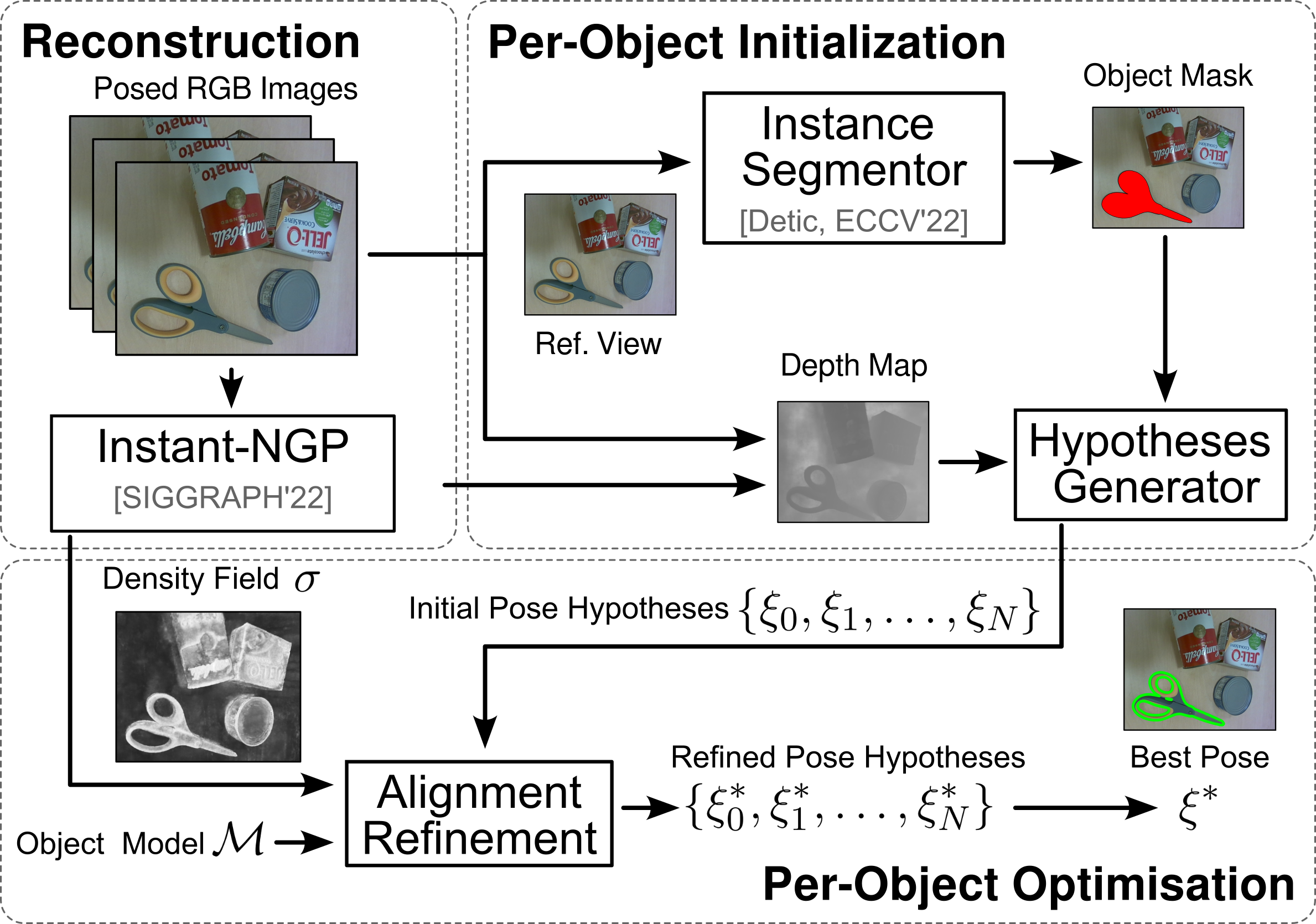}
    \caption{Overview of the proposed framework: an Instant-NGP reconstruction is obtained from images captured from a robot's wrist-mounted camera. 
    Objects of interest are segmented from a reference view, and a depth map rendering from the same view is used to initialise a set of per-object pose hypotheses.
    Each hypothesis is optimised finding the best pose alignment using the Instant-NGP's density field.}
    \label{fig:pipeline}
    \vspace{-0.5cm}
\end{figure}

The core of our method is first to use the posed images to globally reconstruct the density and radiance fields of the scene. 
We then use segmentation in a single view to propose and initialise 3D object model pose hypotheses, which are refined by alignment with the density field reconstruction. 
In this paper we rely on Instant-NGP~\cite{instant_ngp} for radiance field and density field reconstruction, as well as refinement of the original camera poses, which is crucial for accurate reconstruction and model fitting.
In principle, any existing or future radiance and density reconstruction method could be used instead, though it would need to improve on Instant-NGP's remarkable accuracy and efficiency for that to be worth it.
%Combining both the \ac{NeRF} reconstruction and the segmented object's 2D masks, each object is assigned a set of initial per-object pose hypotheses, which are further refined geometrically for the best model-to-scene fitting pose.
% for the alignment of the object model to the scene.
% Each of these hypotheses is further optimized for its alignment against the \ac{NeRF} reconstruction until convergence, and the best fit is identified as the optimal object model-to-scene alignment pose.
An overview of the method is depicted in \reffig{fig:pipeline}.

\subsection{Object Model Representation}

We represent object models $\model$ flexibly as a set of surface points with normals  $(\point_i, \normal_i)$, $\point_i \in \surface\subset\Rthree$, $\normal_i \in \normalset\subset\Rthree$. This representation can be used for object models acquired or designed in different ways.
For instance, in \refsec{sec:experimental_evaluation}, we show the algorithm applied to both human-designed \ac{CAD} models and models built by 3D reconstruction.
Our method tackles object pose alignment in purely geometrical terms, 
and our object models do not need any appearance information, granting the method robustness to varying lighting conditions.
If necessary we  pre-process models with high numbers of points and normals by uniformly sampling $N_\model$ samples across the surface for efficiency.

\subsection{Density fields from Instant-NGP}

We align the object models against a 3D density field $\densityf(\point)$, which maps every 3D scene location $\point \in \Rthree$ to its density or probability of being occupied. 
In our current system, this field comes from Instant-NGP, which optimises a density and radiance field against a set of captured images via volumetric rendering. 
We initialise the camera poses from robot arm kinematics, so no additional camera tracking system such as sparse visual SLAM/structure from motion is needed.

%Here, however, we focus on extracting such a density field from \acp{NeRF}\cite{nerf}, which, in its most basic form, only  requires  a set of posed RGB images to operate. 
%This data is especially simple to acquire in a robotic manipulation setup as the one explored in this paper, where the pose of an image taken from a wrist-mounted camera can be straightforwardly retrieved using forward kinematics.
%In \acp{NeRF}, these posed images supervise the optimisation of a plenoptic function via differential rendering, where the reconstructed density field $\densityf$ is key. % in the synthesization of images.
% The main advantage of \acp{NeRF} over alternatives is that they can seamlessly be applied to complex scenes while handling complex light-scene interactions (e.g., specular reflections, transparencies) or even refining camera-related inaccuracies (e.g., poses, camera calibration) without carefully tuning.
% While initial approaches \cite{nerf,nerfpp} were  prohibitively inefficient for online operation, recent works have been proved capable of such representations \cite{imap, niceslam,instantngp} in near real-time, unlocking their application to interesting robotic applications.

Instant-NGP does not explicitly aim for a high-fidelity density field and thus the quality of the retrieved 3D reconstruction is generally variable. 
However, we will demonstrate that it is sufficient to achieve millimetre-accuracy  object pose estimation without any special post-processing. We only need to capture images and optimise Instant-NGP once per scene, and then multiple objects can be fitted to the same density reconstruction.

%The training of the \ac{NeRF} scene representation is obtained as a pre-processing step in our pipeline, remaining frozen throughout the next stages where multiple objects are identified and fitted to such a reconstruction.

%Here, however, we focus on extracting such a density field using \ac{NeRF} so that the only minimal requirement is to use RGB images $\image \in \R^{w\times h}$ taken from provided poses $\pose \in \SEthree$. 
%In essence, \ac{NeRF} $\nerf_{\param_\nerf}$ combines both a densitiy field with a radiance field, $\colorf(\point, \viewpoint; \param_c)$ that produces a color for each each point $\point \in \Rthree$ observed from a specific viewpoint direction $\viewpoint \in \Rtwo$.
%Via differentiable rendering,  for any given camera pose $\pose$ an image can be synthesized $\hat{\image}(\pose; \Theta_\nerf)$ and supervised against ground-truth images in order to estimate the parameters $\param_\nerf = \{\param_\densityf, \param_\colorf \}$ and even potentially refining the camera poses $\pose$ in some of the state-of-the-art pipelines.
%While our method is mainly agnostic to the employed density field, here we exhaustively research for RGB-only, \ac{NeRF}-derived density fields which only impose some lax requirements for real-world setups.  

\subsection{Multi-hypothesis Object-Pose Optimisation}\label{sec:initialization}

Each object is allocated a fixed number of hypotheses of potential model-to-scene alignment poses $\hypothesesset = \left\{\hypothesis_0,\hypothesis_1,\dots,\hypothesis_{N_\hypothesesset}\right\}, \pose = \{R,\mathbf{p}\}\in \langle\SOthree\times\Rthree\rangle$, that map coordinates in the canonical space of the object onto the scene coordinate system as  $\xi(\point) = R \point +\mathbf{p}, \forall \point \in \Rthree$. 
% , which will be concurrently optimised against the scene density field $\densityf$ (see \refsec{sec:alignment})

\subsubsection{Initialization}
To initialize these per-object hypotheses, we consider a specific view among all the posed images to be the reference view of the scene (e.g., the first captured image, the most top-down view) and apply an off-the-shelf 2D instance segmentor \cite{detic} to identify all the relevant objects in the scene
for which we have 3D models.
We render a depth map from Instant-NGP for this reference view and extract a partial 3D point cloud from the 2D pixel mask of each identified object. We then generate multiple 3D pose hypotheses $\hypothesesset$, using
the centre of mass of the partial point cloud as the translation of every hypothesis and
rotations are equally distributed in $\SOthree$.
%This procedure implements a simple initialisation approach, which could be further complemented by using, for instance, multi-view point cloud merging or 2D single-view pose estimation\cite{1shotpose} or 2D multi-view pose estimation \cite{cosypose}.
In \refsec{sec:experimental_evaluation} we show that this simple initialization is sufficient to achieve good model fitting in real-world experiments.

\subsubsection{Alignment Refinement}\label{sec:alignment}
% For each object model $\model$, we define a set of points $\pointsurfaceset$ closely distributed around the surface of the model $\pointsurfaceset = \left\{\pointsurface \;|\; \pointsurface =  \point_i + \normal_i\left[-d^\surface,d^\surface\right], \forall \left\{\point_i,\normal_i\right\} \in \left\{\surface, \normalset \right\}\right\}$, and a set of points along the surface normals $\pointnormalset = \left\{\pointnormal \;|\; \pointnormal=  \point_i + \normal_i\left[d^\surface, d^\surface+d^\normalset]\right], \forall \left\{\point_i,\normal_i\right\} \in \left\{\surface, \normalset \right\}\right\}$

For each object model $\model$, we define a set of points $\pointsurfaceset$ closely distributed around the surface of the model and a set of points distributed outwards along the surface normals $\pointnormalset$.
Formally:
\begin{align}
\pointsurfaceset &= \left\{\pointsurface \;|\; \pointsurface =  \point_i + \normal_i\left[-\delta^\surface,\delta^\surface\right]\right\}\\
\pointnormalset &= \left\{\pointnormal \;|\; \pointnormal=  \point_i + \normal_i\left(\delta^\surface, \delta^\surface+\delta^\normalset\right]\right\}
\end{align}
$\forall \left\{\point_i,\normal_i\right\} \in \left\{\surface, \normalset \right\}$, with $\delta^\surface, \delta^\normalset \in \R^+$  parametrising the uniform sampling in the defined intervals along the normals.

\begin{figure}
    \centering
    \includegraphics[width=0.45\textwidth]{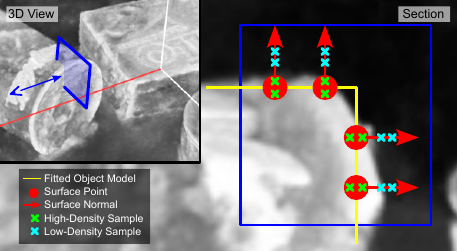}
    \caption{Section of a reconstructed density field within object.
    Note that, while the RGB renders from a \ac{NeRF} can achieve high-fidelity, the underlying density field can be as noise as shown here, even after \ac{NeRF} convergence.
    This noisy density field is used in our fitness function \refeq{eq:alignment}, promoting that points near the surface of the aligned model object $\pointsurfaceset$ and points along their normal $\pointnormalset$  fall within high-density or low-density region, respectively.}
    \label{fig:section_density}
    \vspace{-0.5cm}
\end{figure}

For a good model-to-scene fit, all the points near the surface of the model $\pointsurfaceset$ should be projected to high-density regions of the density field $\densityf$ whereas the points projected outwards along the normals $\pointnormalset$, away from the surface into free space,  should be mapped to low-density regions.
We retrieve an occupancy measurement from Instant-NGP's density field as $\sample(\point) = 1-\exp(-\exp(\densityf(\point))\beta), \point\in\Rthree$, which resembles the expression for the light transmittance used in differential rendering using a tuneable parameter $\beta\in\R^+$. 
As Instant-NGP's density field is often quite irregular (see \reffig{fig:section_density}), $\delta^\surface $ and $\delta^\normalset$ help in creating bands of points around the surface or out of it, i.e. $\pointsurfaceset$ or  $\pointnormalset$, which are largely expected to be occupied or empty, respectively, for the best model-to-scene fit. 
Formally, we define the model-to-scene fitness function:
\begin{align}
    \fitnessfunc(\hypothesis) = \frac{1}{\left|\pointsurfaceset\right|}\sum_{\pointsurface \in \pointsurfaceset} s(\hypothesis(\pointsurface)) - \frac{1}{\left|\pointnormalset\right|}\sum_{\pointnormal \in \pointnormalset} s(\hypothesis(\pointnormal)),\label{eq:alignment}
\end{align}
which can be used to convert all the initial pose hypotheses $\left\{\hypothesis_0,\hypothesis_1,\dots,\hypothesis_{N_\hypothesesset}\right\}$ into refined ones $\left\{\hypothesis_0^*,\hypothesis_1^*,\dots,\hypothesis_{N_\hypothesesset}^*\right\}$ via non-linear optimisation. 
Among all the refined pose hypotheses, the one that fits the best \refeq{eq:alignment} is selected as the optimal pose of the object.
Depending on the number of hypotheses per object $N_\hypothesesset$ and the complexity of the geometry, it is not uncommon that multiple initial hypotheses are refined to the same final pose. In practice, each hypothesis is independent of the others and thus their refinement is executed in parallel as a batch.

\section{Experimental Evaluation}
\label{sec:experimental_evaluation}

\begin{figure*}
    \centering

    % Experiment 13 results
    \begin{subfigure}[b]{0.18\textwidth}
    \includegraphics[angle=180, origin=c, width=\linewidth]{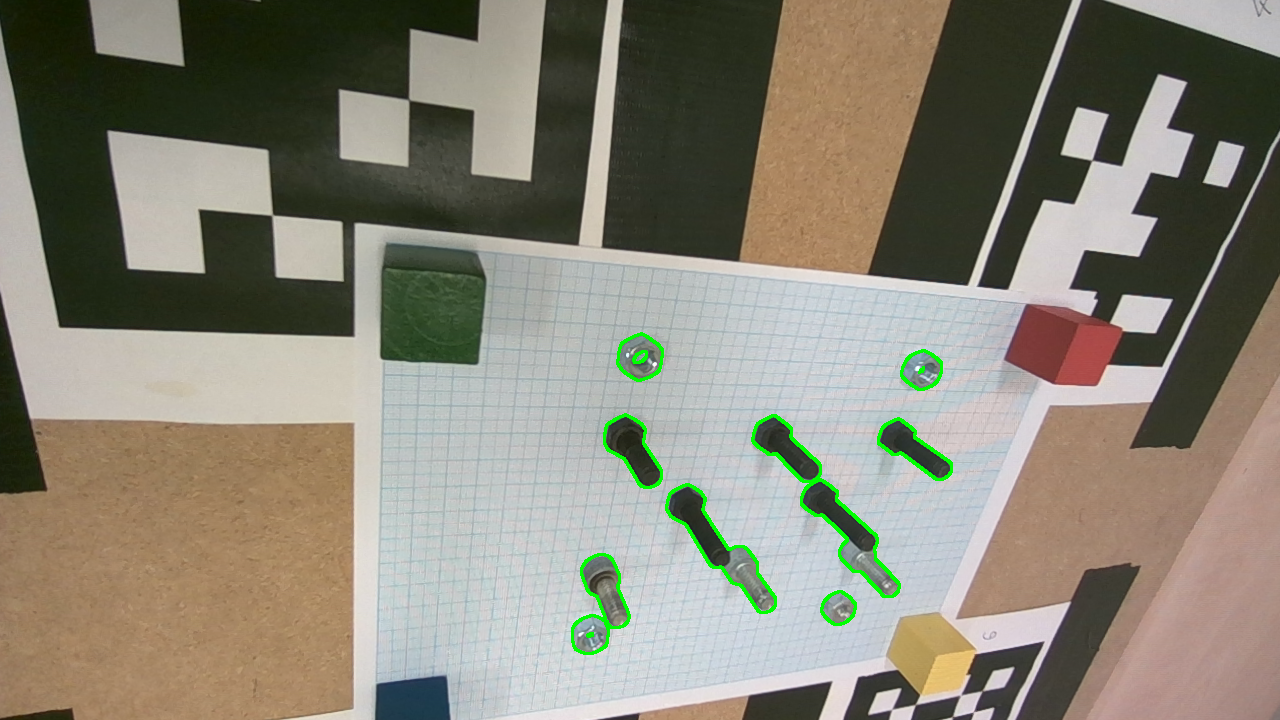}
    \end{subfigure}    
    \begin{subfigure}[b]{0.18\textwidth}
    \includegraphics[angle=180, origin=c, width=\linewidth]{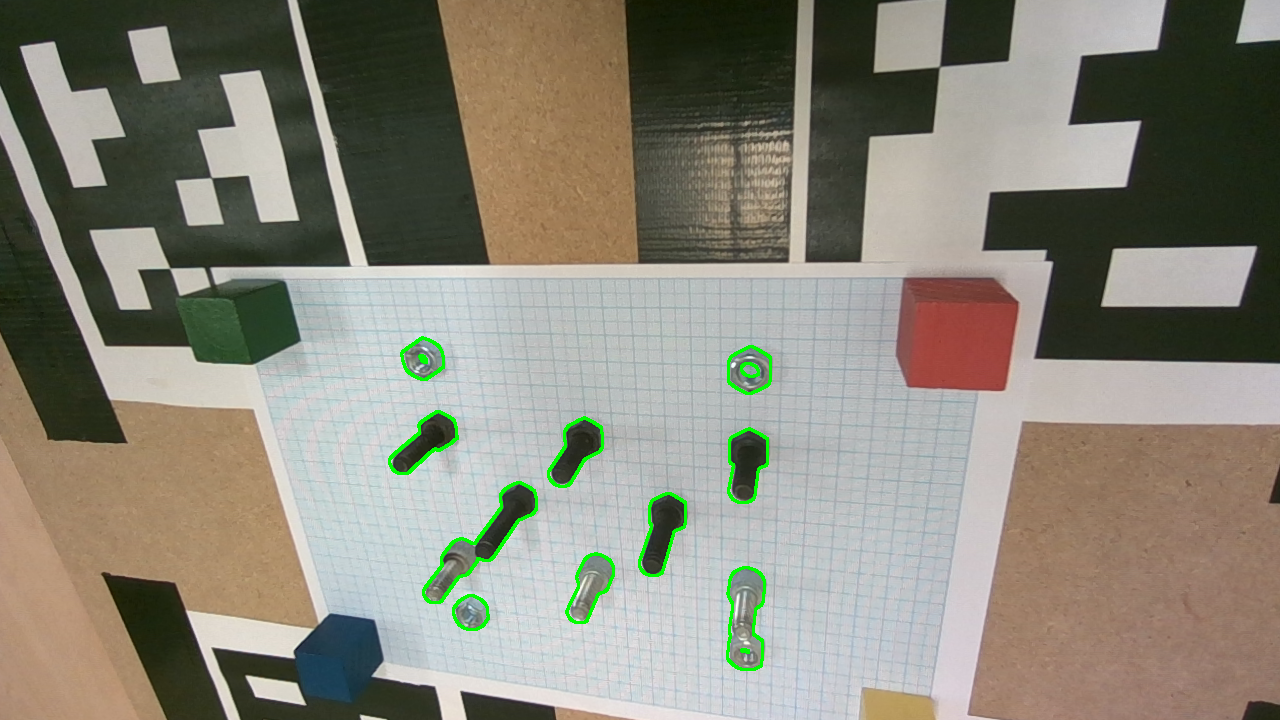}
    \end{subfigure}    
    \begin{subfigure}[b]{0.18\textwidth}
    \includegraphics[angle=180, origin=c, width=\linewidth]{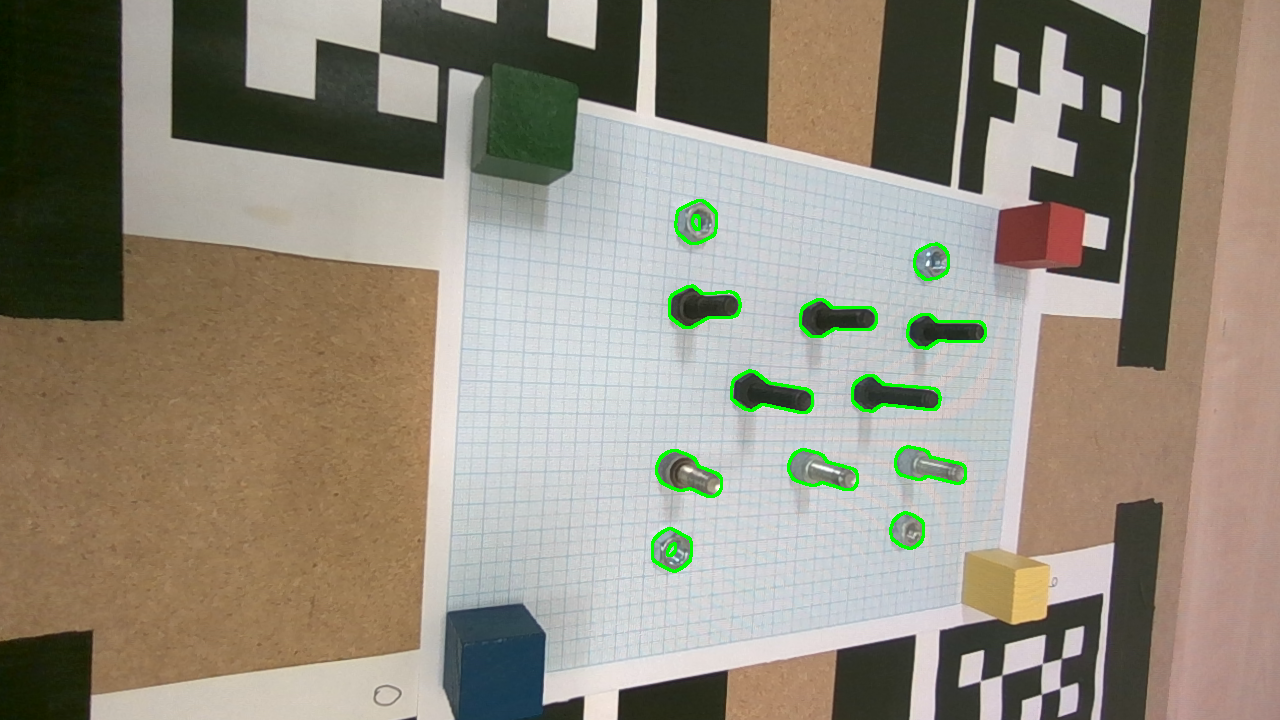}
    \end{subfigure}
    \begin{subfigure}[b]{0.18\textwidth}
    \includegraphics[angle=180, origin=c, width=\linewidth]{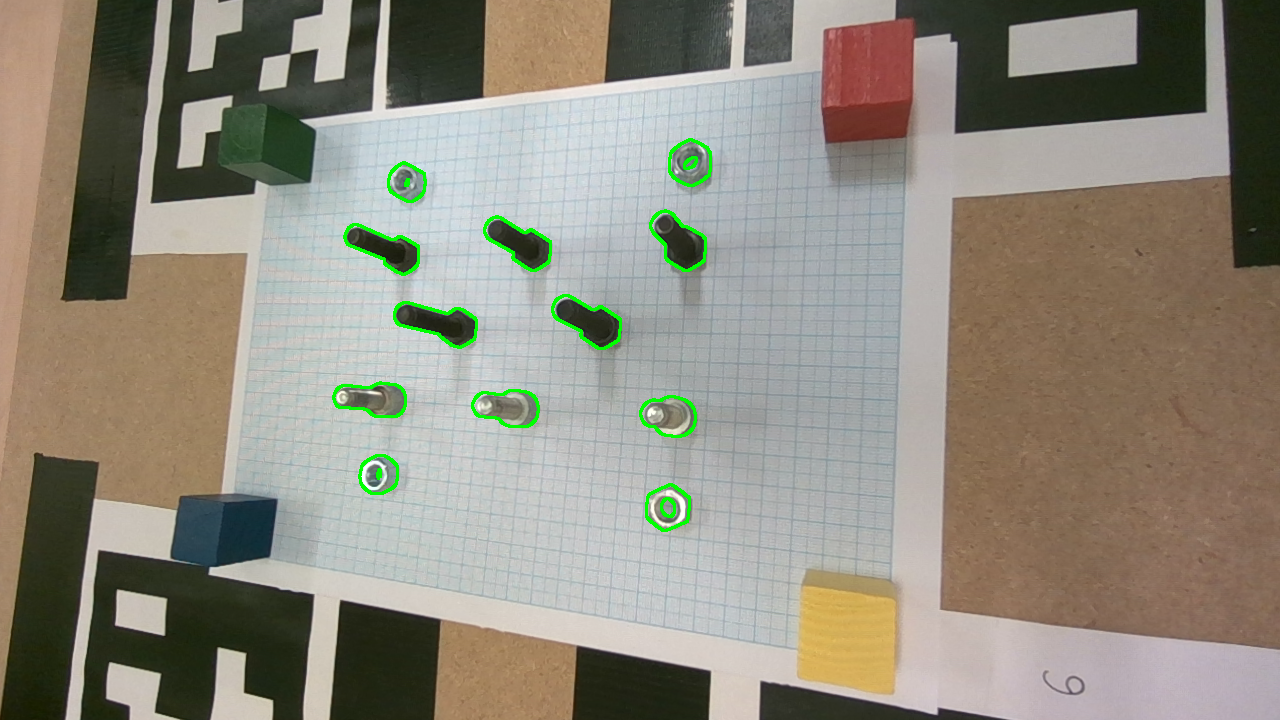}
    \end{subfigure}
    \begin{subfigure}[b]{0.18\textwidth}
    \includegraphics[angle=180, origin=c, width=\linewidth]{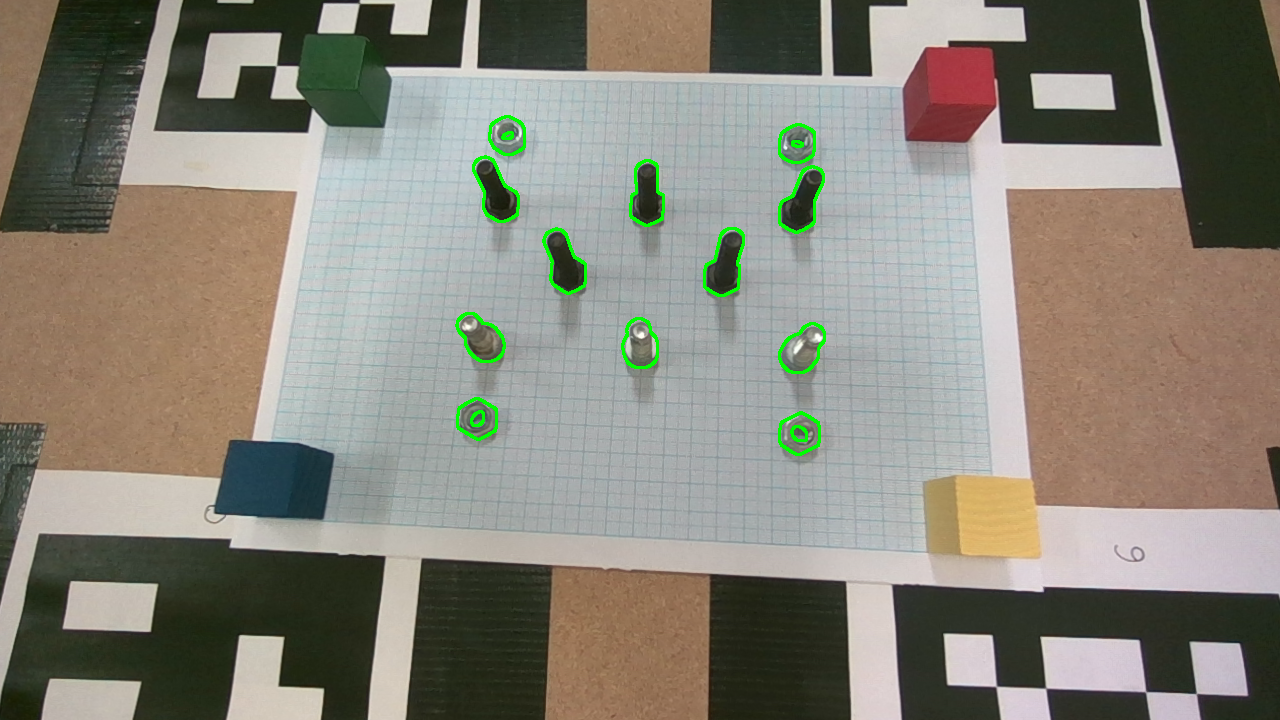}
    \end{subfigure}

    % Experiment 0 results
    \begin{subfigure}[b]{0.18\textwidth}
    \includegraphics[angle=180, origin=c, width=\linewidth]{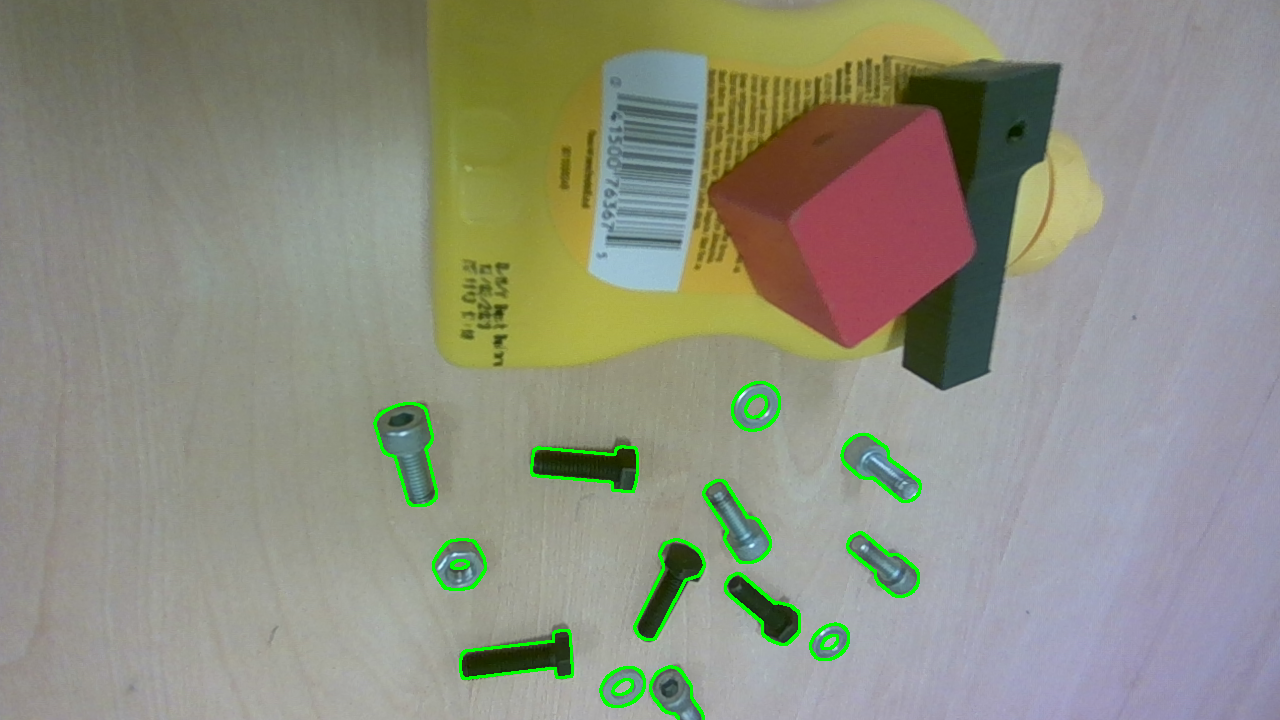}
    \end{subfigure}    
    \begin{subfigure}[b]{0.18\textwidth}
    \includegraphics[angle=180, origin=c, width=\linewidth]{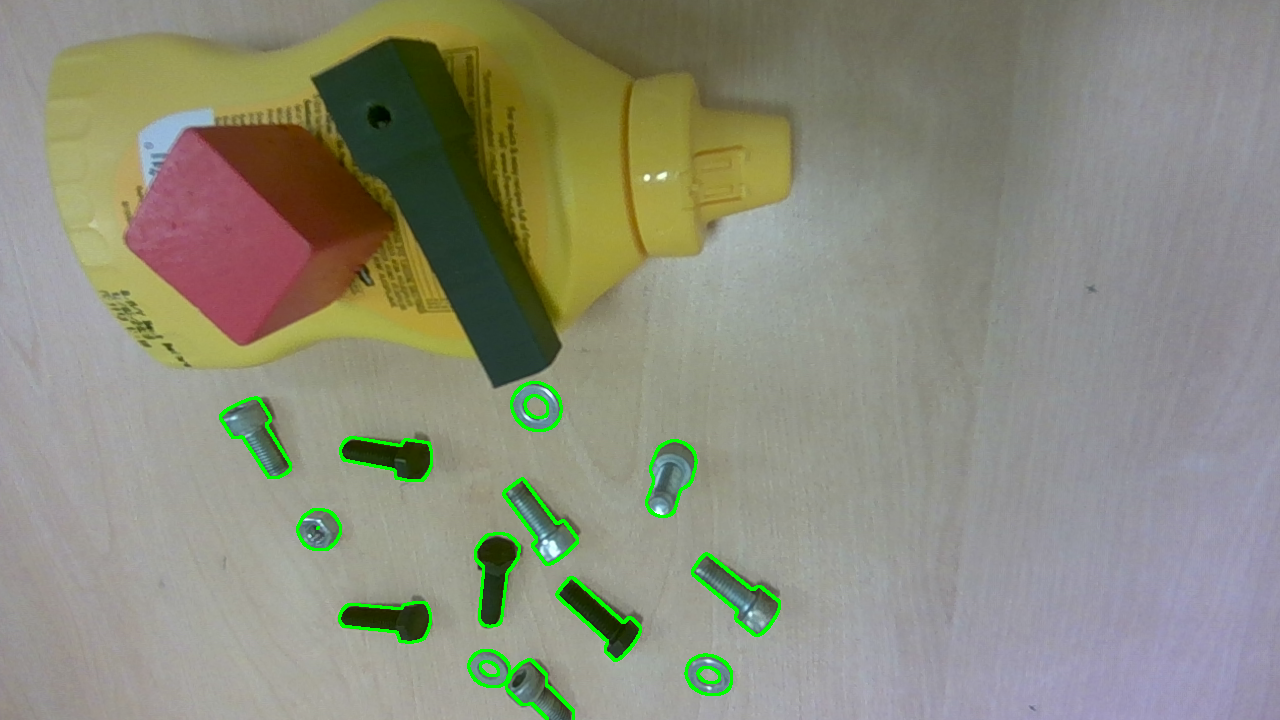}
    \end{subfigure}    
    \begin{subfigure}[b]{0.18\textwidth}
    \includegraphics[angle=180, origin=c, width=\linewidth]{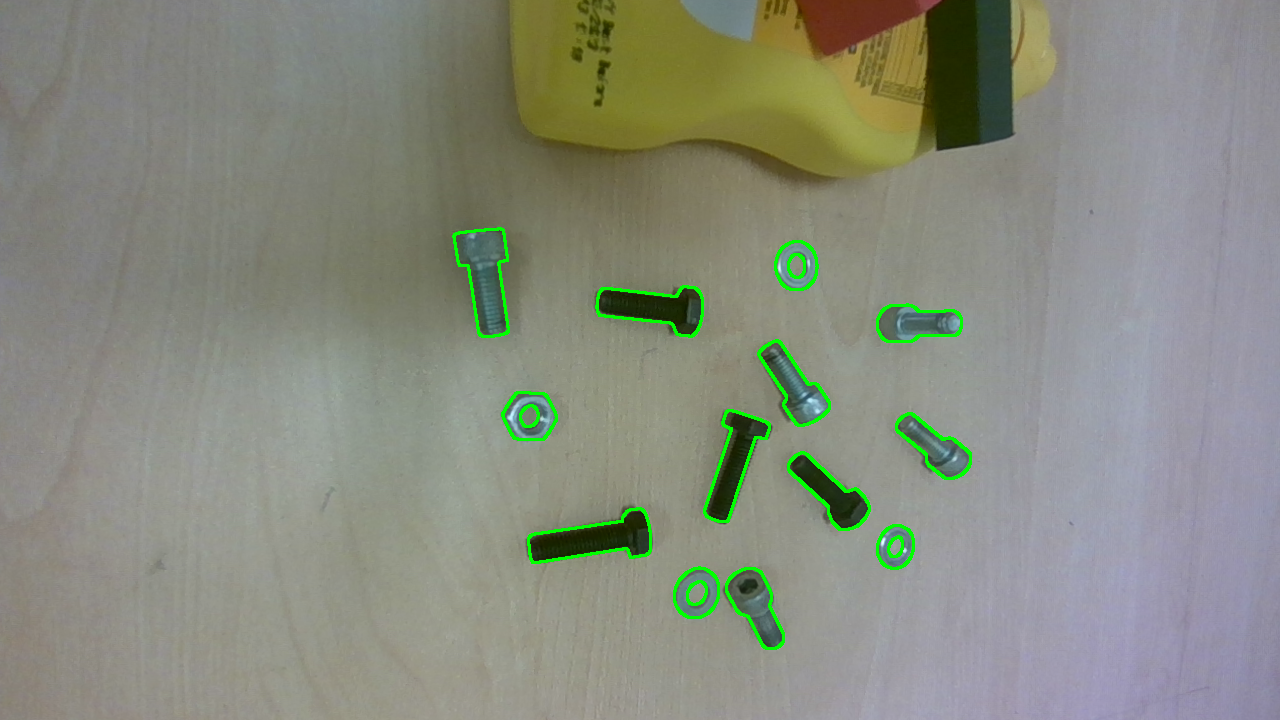}
    \end{subfigure}
    \begin{subfigure}[b]{0.18\textwidth}
    \includegraphics[angle=180, origin=c, width=\linewidth]{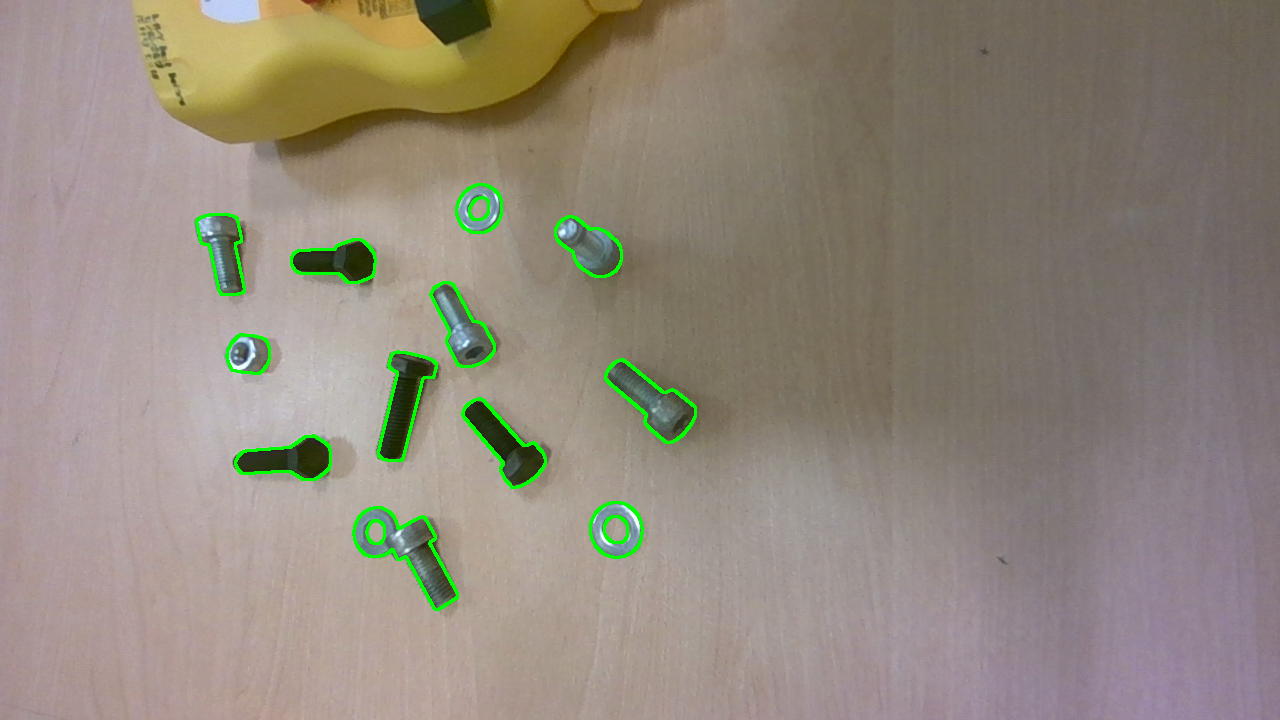}
    \end{subfigure}
    \begin{subfigure}[b]{0.18\textwidth}
    \includegraphics[angle=180, origin=c, width=\linewidth]{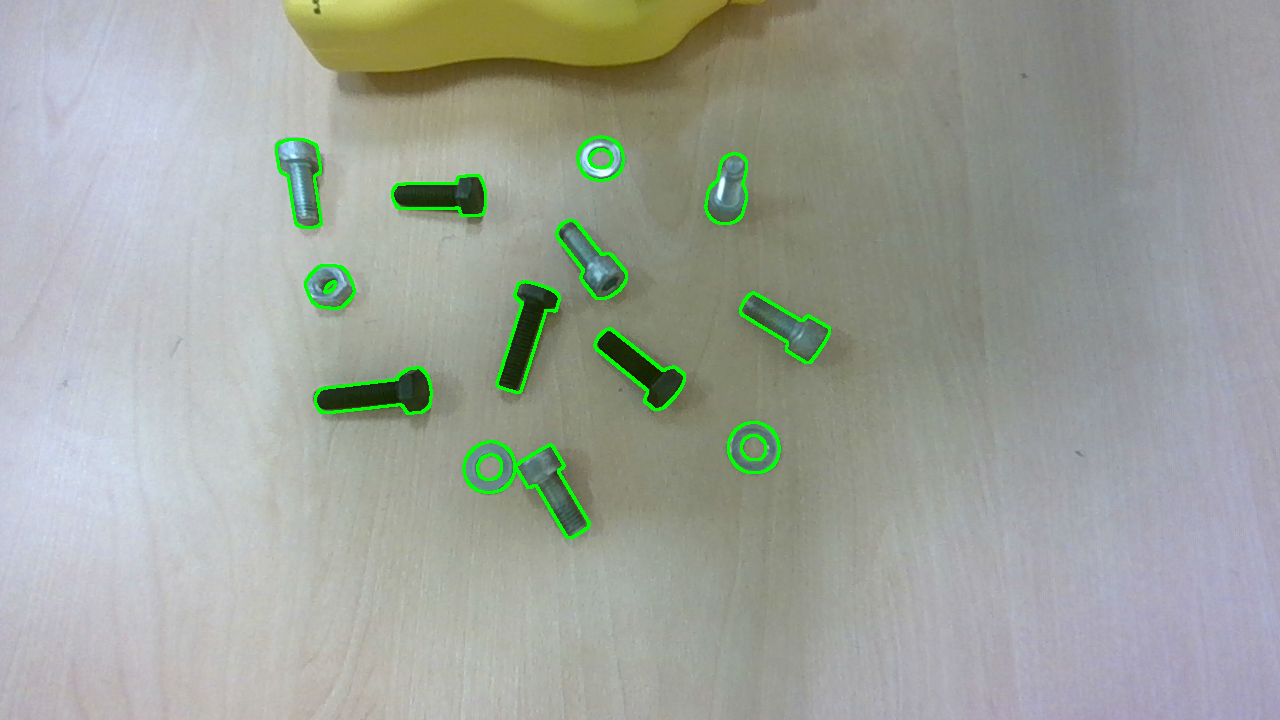}
    \end{subfigure}

    % Experiment 6 results
    \begin{subfigure}[b]{0.18\textwidth}
    \includegraphics[angle=180, origin=c, width=\linewidth]{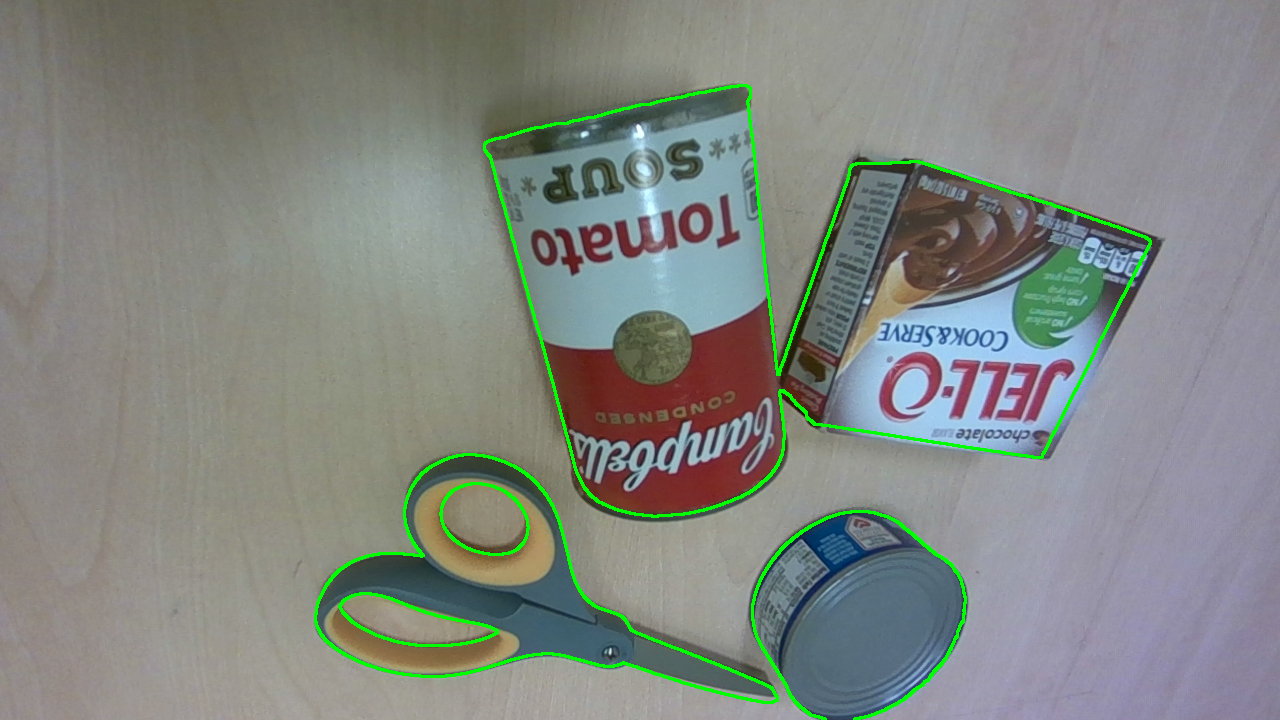}
    \end{subfigure}    
    \begin{subfigure}[b]{0.18\textwidth}
    \includegraphics[angle=180, origin=c, width=\linewidth]{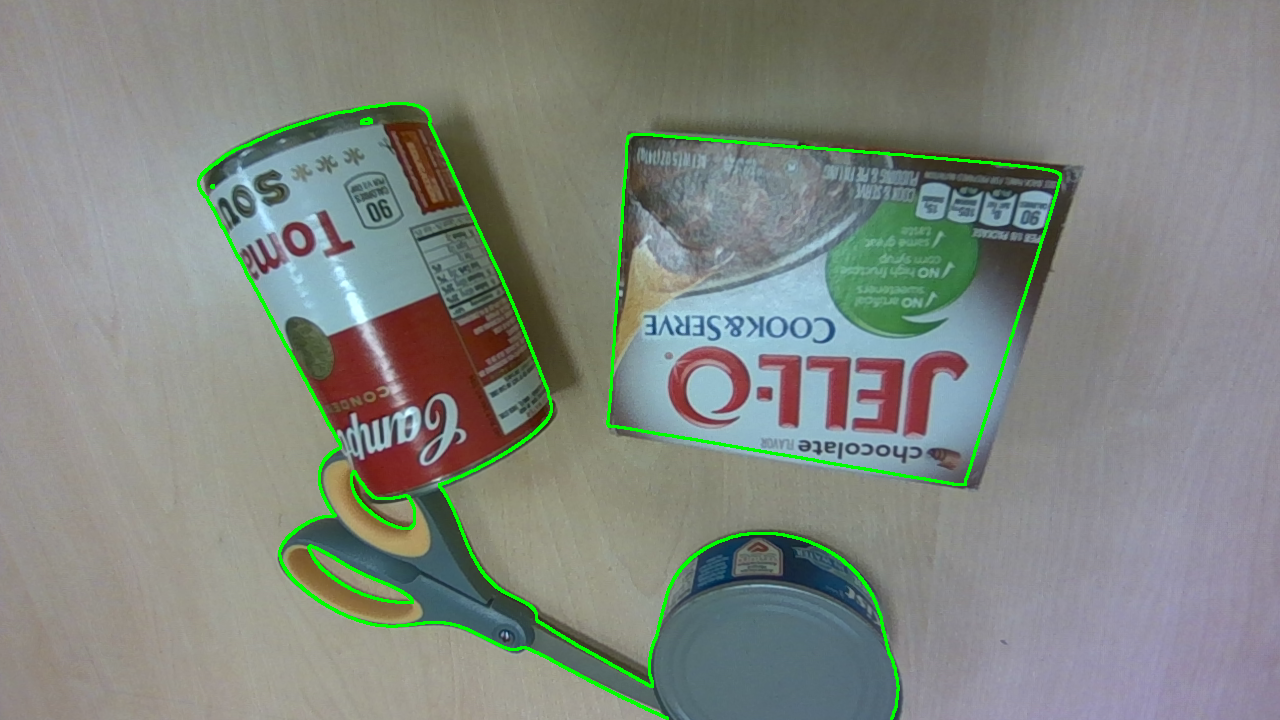}
    \end{subfigure}    
    \begin{subfigure}[b]{0.18\textwidth}
    \includegraphics[angle=180, origin=c, width=\linewidth]{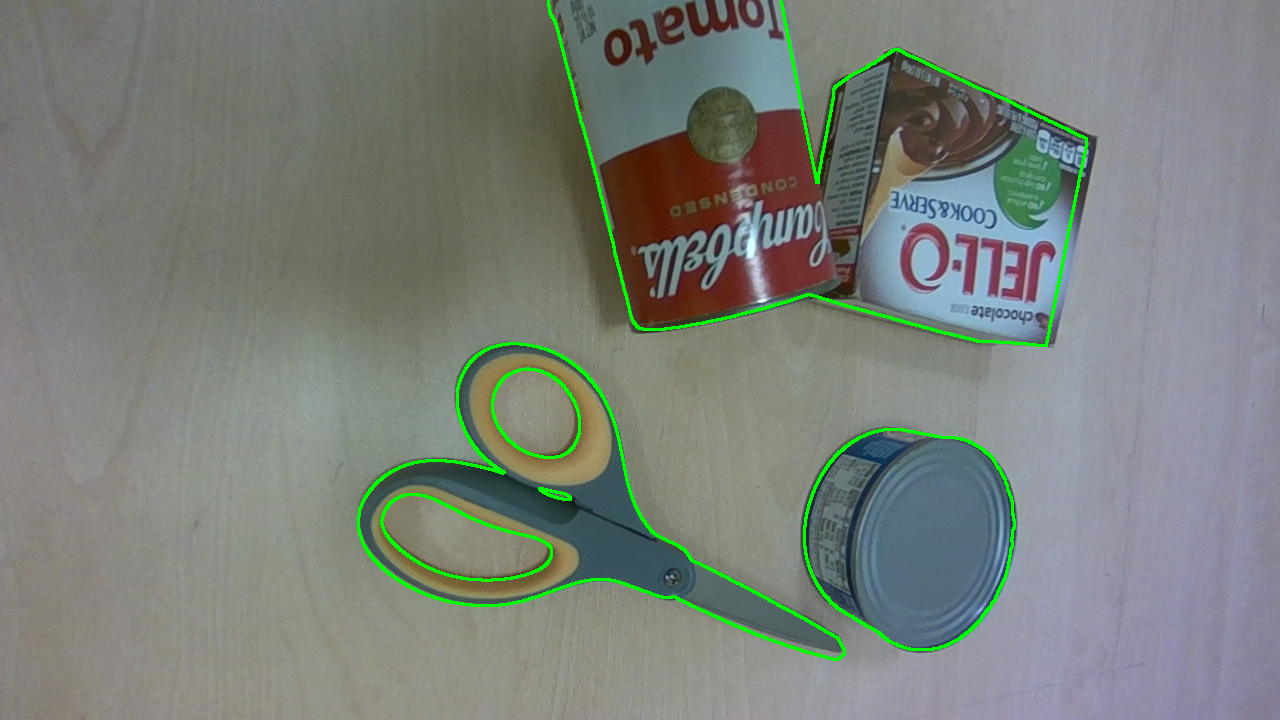}
    \end{subfigure}
    \begin{subfigure}[b]{0.18\textwidth}
    \includegraphics[angle=180, origin=c, width=\linewidth]{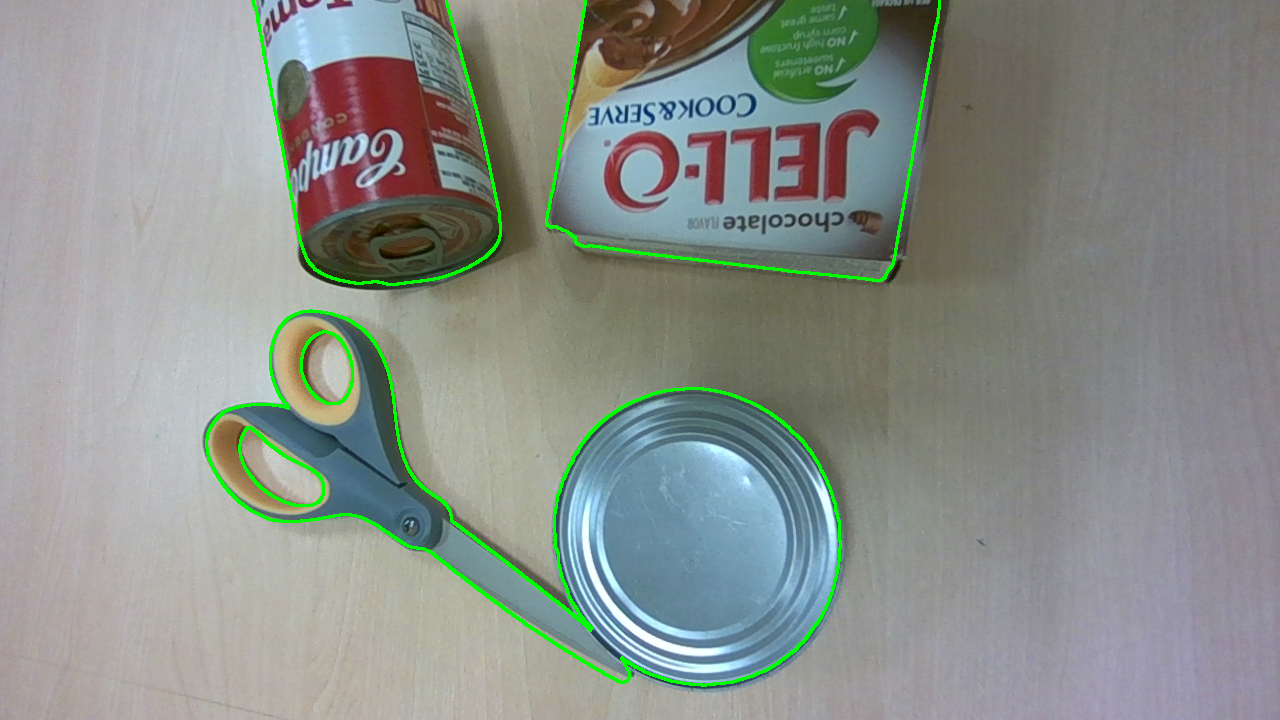}
    \end{subfigure}
    \begin{subfigure}[b]{0.18\textwidth}
    \includegraphics[angle=180, origin=c, width=\linewidth]{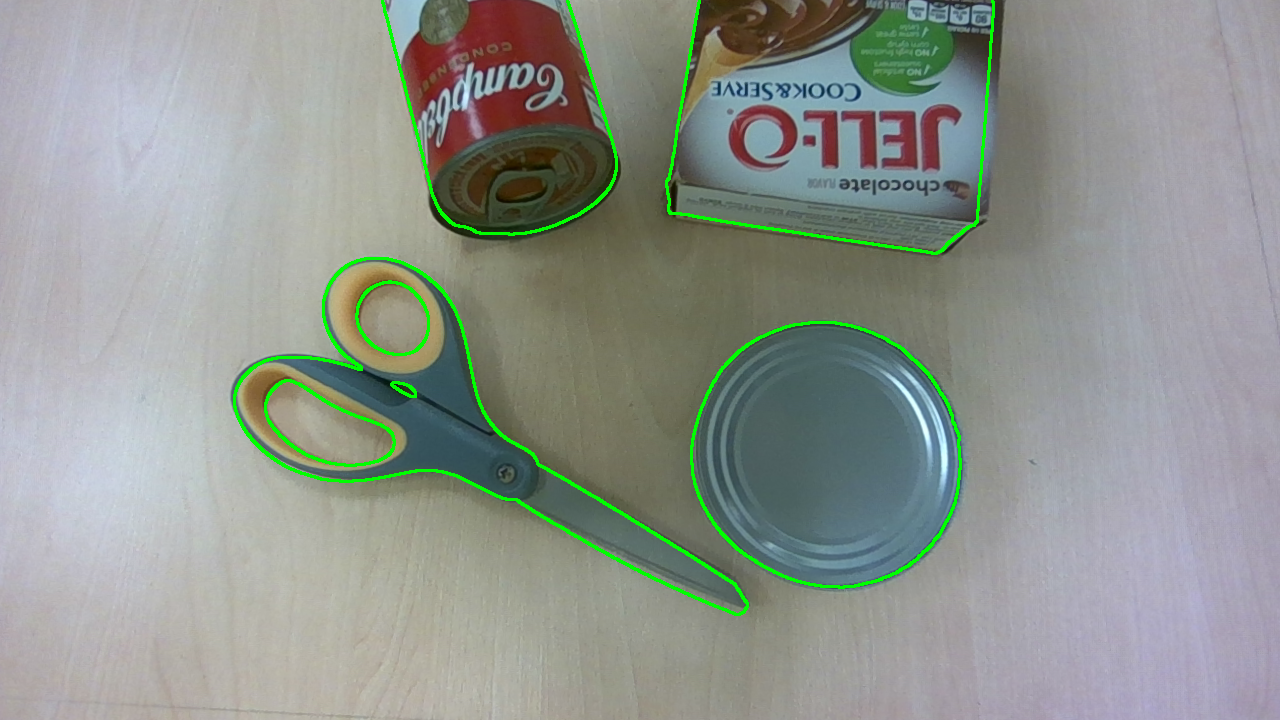}
    \end{subfigure}

    % Experiment 8 results
    \begin{subfigure}[b]{0.18\textwidth}
    \includegraphics[angle=180, origin=c, width=\linewidth]{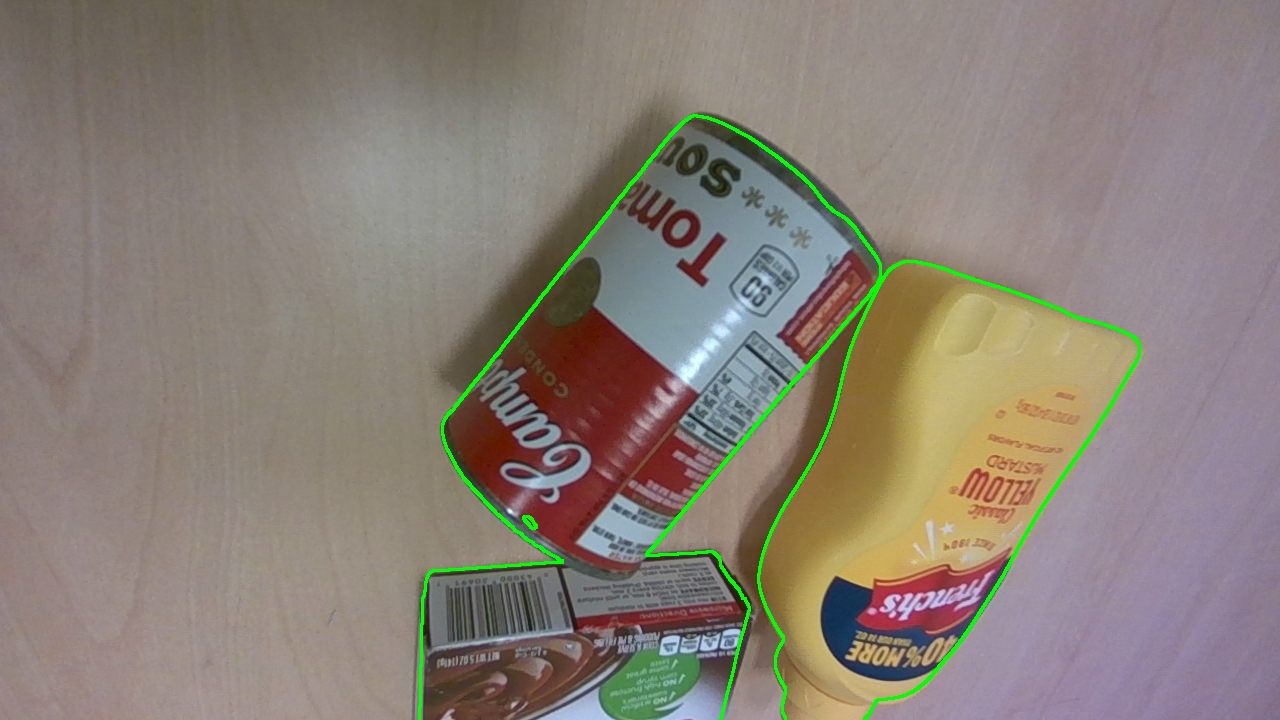}
    \end{subfigure}    
    \begin{subfigure}[b]{0.18\textwidth}
    \includegraphics[angle=180, origin=c, width=\linewidth]{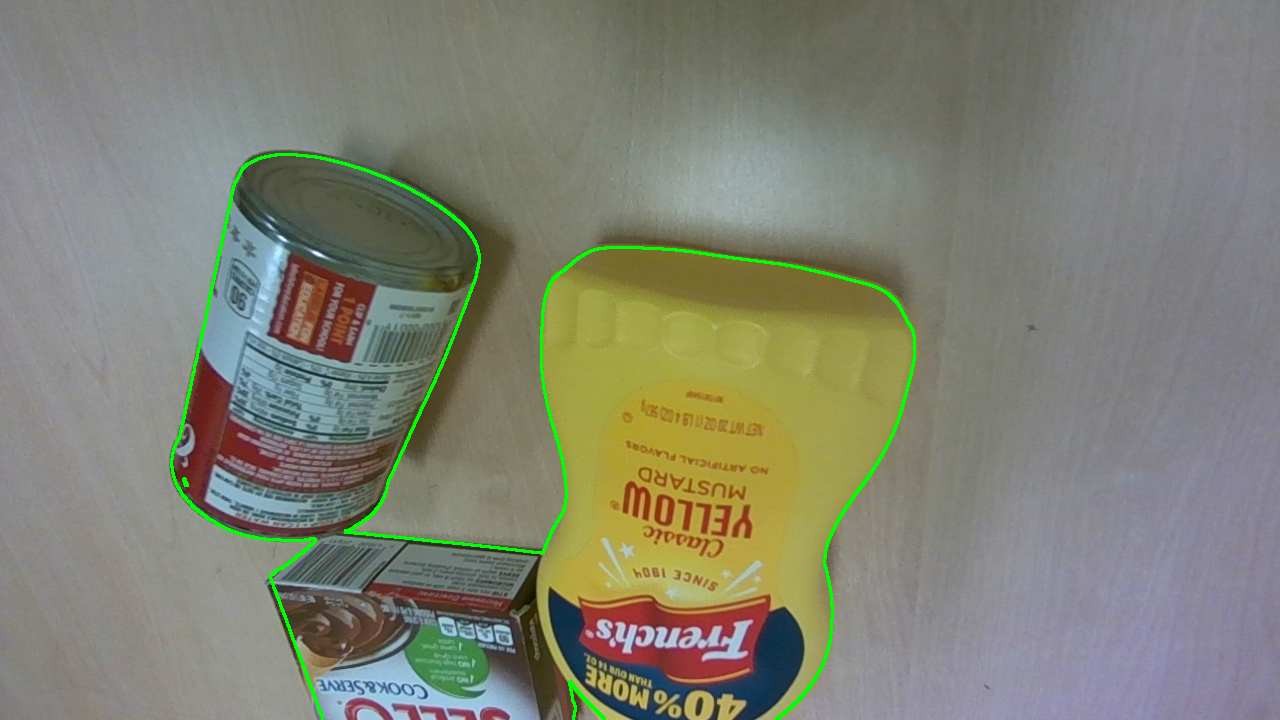}
    \end{subfigure}    
    \begin{subfigure}[b]{0.18\textwidth}
    \includegraphics[angle=180, origin=c, width=\linewidth]{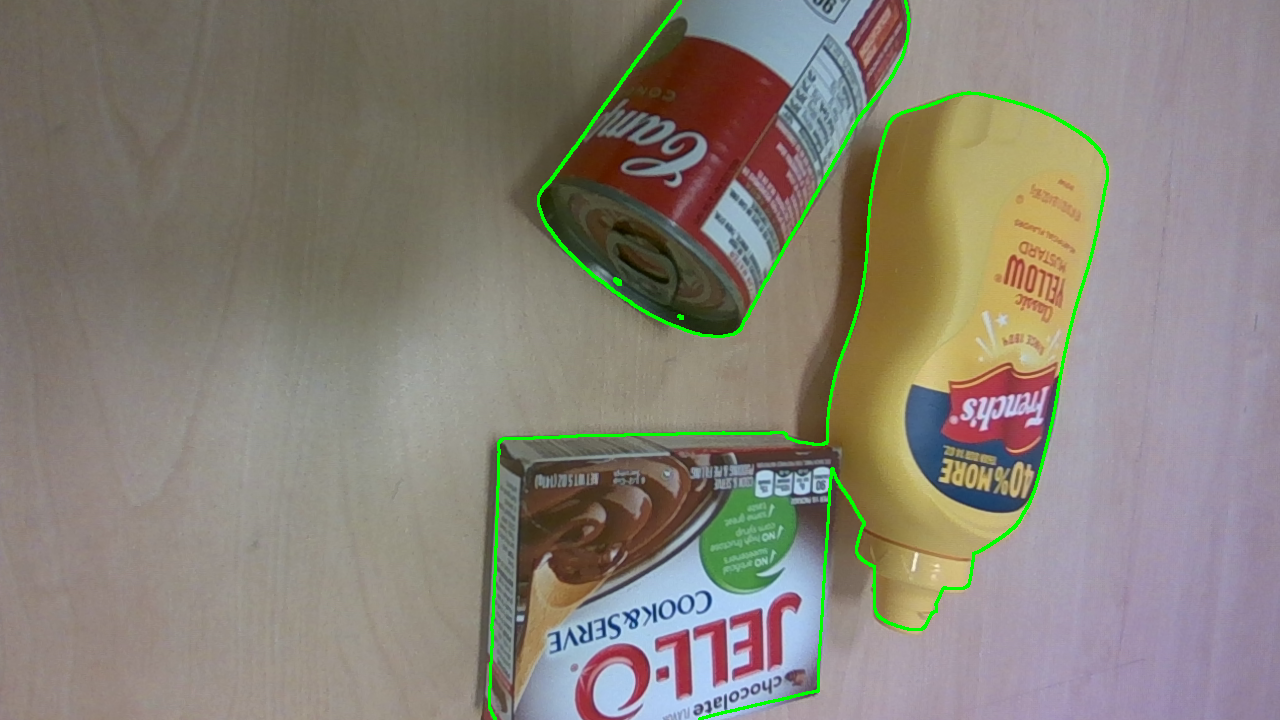}
    \end{subfigure}
    \begin{subfigure}[b]{0.18\textwidth}
    \includegraphics[angle=180, origin=c, width=\linewidth]{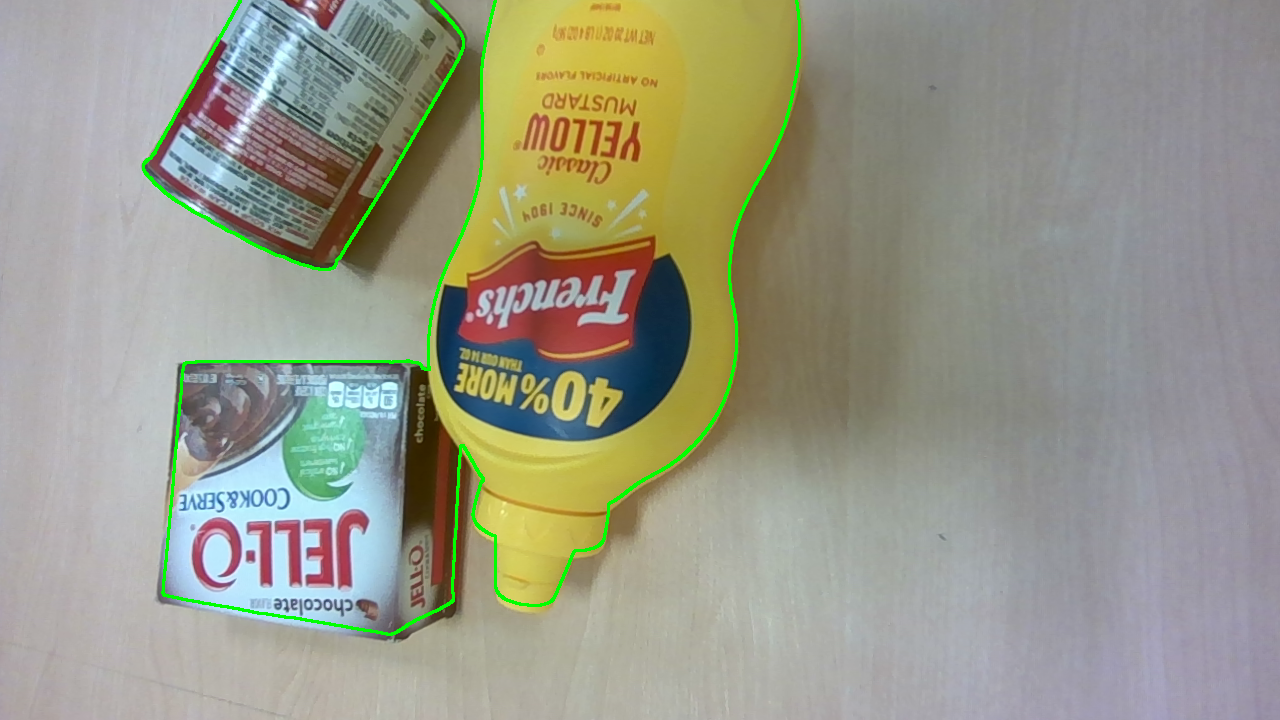}
    \end{subfigure}
    \begin{subfigure}[b]{0.18\textwidth}
    \includegraphics[angle=180, origin=c, width=\linewidth]{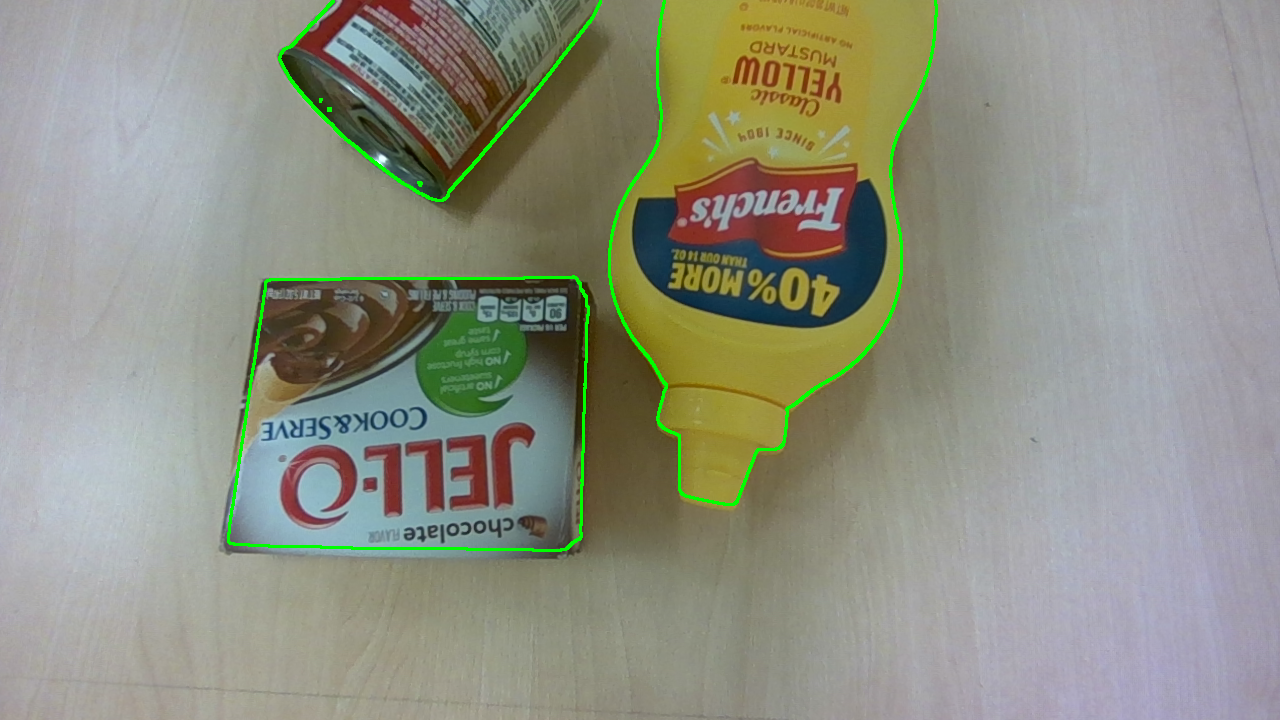}
    \end{subfigure}

    \caption{Self-collected real-world datasets where the poses of all the relevant objects are accurately estimated, as evidenced by the re-projection of the silhouettes of their models perfectly aligning in the image plane.}
    \label{fig:fitting_demos}
\end{figure*}

We now showcase the capabilities of our method in a challenging robotic experimental setup. 
%This evaluation assesses the quality of the retrieved object poses while providing insights of the key elements of the pipeline and their impact on the overall system performance.
We explore a series of self-collected experiments with multiple objects loosely arranged in a table-top configuration (see \reffig{fig:fitting_demos}).
Many of the objects used are small, have complex geometries or would require tight tolerances in their manipulation, justifying the need for highly accurate pose estimation.
Additionally, we focus on objects that are visually challenging for other techniques, with textureless or reflective surfaces.

\subsection{Implementation Details and Experimental Setup}
%Our framework employs Instant-NGP \cite{instantngp} as our \ac{NeRF} backbone as its fast convergence and good reconstruction capabilities enable the deployment of our pipeline for realistic and efficient downstream robotic applications.

In our experiments, the scene is observed using a Franka Emika robot arm equipped with a wrist-mounted Real-sense Sense D455 1280×720 camera (only RGB data is used), interfaced via  ROS \cite{ros} and hand-eye calibrated as per Tsai \etal~\cite{Tsai_calibration}.
For each experiment, we automatically execute a precomputed hemispherical trajectory, capturing up to 78 posed images pointing towards the centre of the scene which are used to train Instant-NGP. 
While the whole scanning process takes over a minute, it might be potentially too slow for some potential downstream applications.
Therefore, in our experiments, we investigate how the performance of the system would be impacted in a less complete albeit faster scanning (see \refsec{sec:experiment:ablation}).
Although poses for the captured views are initially retrieved via robot arm kinematics, we have found that potential inaccuracies (e.g., arm encoder noise, deviations in the hand-eye calibration) are handled by enabling the camera poses refinement in Instant-NGP, yielding higher fidelity density fields.
Instant-NGP is trained once per scene and used in frozen form for pose estimation of all of the objects.

Once Instant-NGP has been trained, we employ Detic \cite{detic} to create 2D instance object masks for hypothesis initialization as described in \refsec{sec:initialization}.
Each object is allocated $N_\hypothesesset=216$ hypotheses, refined by optimising \refeq{eq:alignment} via non-linear optimisation for $200$ iterations using Adam \cite{adam} in Pytorch \cite{pytorch} and LieTorch \cite{lie_torch}, with different learning rates $l_\text{Rot}=2.5\mathrm{e}{-2}$ and $l_\text{Trans}=1.0\mathrm{e}{-3}$  for rotation and translation, respectively. In our experiments, we set $\beta=0.01$.
Note that while some experiments show objects arranged on a flat table-top, our system is not specifically tailored for this case, and always estimates full 6-\ac{DoF} poses.

In our experiments, we consider a mixture of YCB \cite{ycb} and other common objects such as bolts and washers, for which 3D reconstructions and CAD designs are available, respectively.
Each object model $\model$ is uniformly subsampled to only consider a set of $N_\surface=1280$ points along the surface and their normals.
While the definition of $\pointsurfaceset$ and $\pointnormalset$ allows for the generation of many points along each normal according to the intervals defined by $\delta^\surface$ and $\delta^\normalset$, in our experience a single point on the object the surface ($\delta^\surface=0$) and a single point in the direction of the normal at a fixed distance $(\delta^\normalset=5.0\mathrm{e}{-3})$ is sufficient to retrieve accurate poses for simple object geometries, while greatly reducing the computational cost of querying the density field $\densityf$. This is further tested in the ablation study of \refsec{sec:experiment:ablation}.

The whole pipeline is run on a single machine with an NVIDIA Geforce RTX 3090 GPU, i7 Intel CPU and 64GB RAM. 
A single object is optimised in under 3 seconds; this time includes the concurrent optimisation of all $216$ hypotheses. This duration can be significantly reduced by using better initial estimates which can allow us to run with a lower number of hypotheses.

\subsection{Accuracy Evaluation}
\label{sec:experiments:accuracy}
\textbf{\textit{Quantitative Results}.}
To assess the pose estimation accuracy, we collected our own dataset of 4 different scenes each containing a subset of standardised, industrial-grade low-tolerance objects with accurate and widely available CAD models. The dataset in aggregate is comprised of 43 object instances, containing: M8 nuts, M8x25 bolts, M8x30 hex bolts and M8x25 socket bolts.
The small dimensions of these objects render external positioning systems largely inapplicable for highly accurate pose estimates.
Hence, we manually aligned them using technical graph paper with a 1 mm grid division to establish our ground truth.
We purposely placed the objects so that their relative 3D poses can be accurately retrieved from the graph paper up to 1 mm in translation and up to 4 degrees in rotation, given the geometry of the objects.
We compare the estimated relative pose for all object-to-object combinations (considering object symmetries as in \cite{cosypose}) to their ground-truth counterparts.
On this dataset, we report a median relative translation error of 1.6 mm and a median relative rotation error of 3.3 degrees. 
See \reffig{fig:fitting_demos} (top row)  for illustrative examples.

\textbf{\textit{Qualitative Results}.}
We also showcase the capabilities of our system with a diverse set of objects, with scenes composed of YCB objects \cite{ycb} and other common objects in randomly placed configurations. 
While these scenes offer a wider range of interesting configurations compared to the aforementioned dataset, millimetre-accurate ground-truth poses for such arrangements cannot be obtained without the use of specialised setups. Thus, we present here a qualitative evaluation instead, as depicted in \ref{fig:fitting_demos}.
In these scenes, the proposed system is able to accurately retrieve the pose of each of the objects as evidenced by the object silhouettes reprojected onto the image space from their estimated poses.
Note that these scenes include objects that are visually challenging to reconstruct due to, for instance, reflective surfaces (e.g. cans), or small dimensions (e.g. M8 washers), even for Instant-NGP.
Additionally, our system is able to simultaneously operate on objects with very different scales (e.g. note the size difference of the M8 nuts with respect to the mustard bottle).

\begin{figure}
    \centering
    \includegraphics[width=0.42\textwidth]{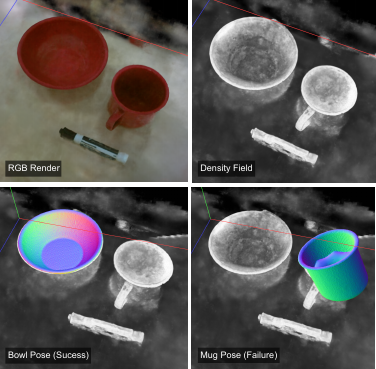}
    \caption{Example of failure case of the proposed system. 
    Instant-NGP is able to produce photorealistic RGB renderings (top left) even when the quality of the underlying density field is poor (top right).
    Despite this, object poses can still be retrieved provided that pose hypothesis initialisation is sufficiently close to the optima (bottom left), resulting in failure otherwise (bottom right).
    }
    \label{fig:failure_case}
\end{figure}

\reffig{fig:failure_case} shows a typical failure case of our algorithm.
While Instant-NGP density field reconstruction is generally poor, in our experience, it is often sufficiently good for object pose estimation but only if the initial set of pose hypotheses are sufficiently close to the global optima.
As our system only implements a simple initialization step based on rendered depth maps, where such a reconstruction is significantly erroneous (e.g. Instant-NGP cannot accurately disambiguate the local geometry), the initial set of poses will be too far from the optima to yield any meaningful pose estimation.

\begin{comment}

so we opt to evaluate our pose estimation, by projecting the CAD models to the Nerf training images and comparing their intersection with segmentation masks obtained using Detic \cite{detic}. For these set of experiments we use CAD models provide in the YCB-Video dataset \cite{posecnn}, since there models are higher in fidelity than the models provided in the YCB dataset \cite{ycb}.

\textbf{Can we do any occlusion experiments??}

In Figure XXX we show our method successfully fitting M8 washers that are only xmm thick and are highly reflective.

\begin{table}[]
\centering
\begin{tabular}{lllll}
\hline
  & Scene A                    & Scene  B &  Scene C & Scene D \\ \hline
\% IOU & \multicolumn{1}{c}{} &   &  &   \\ \hline
\end{tabular}
\caption{Results of projecting the fitted CAD models to the training frames and measuring the IOU with object masks.}
\end{table}
\end{comment}

\subsection{Performance over Varying Views}
\label{sec:experiments:num_views}
\begin{figure}
    \centering
    \includegraphics[width=0.40\textwidth]{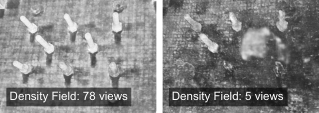}    
    \caption{Density field with varying number of views.}
    \label{fig:df_degrad}
\end{figure}

\begin{figure}
    \centering
    \includegraphics[width=0.48\textwidth]{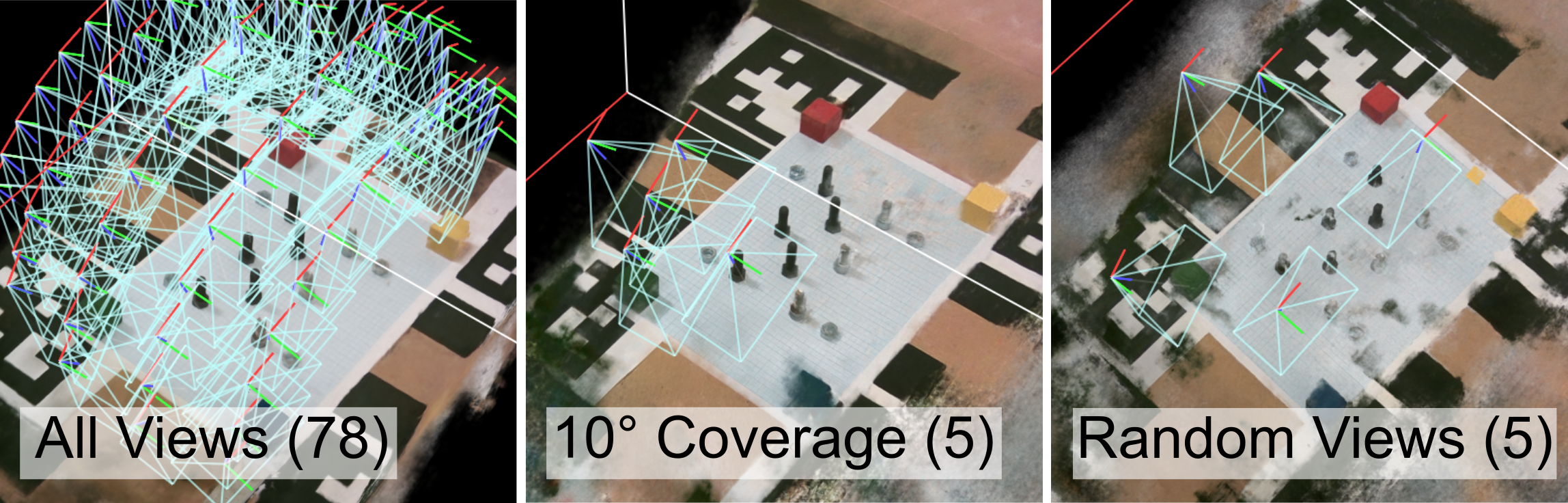}
    \caption{The number of training views (in parentheses) imposes a trade-off between the quality of Instant-NGP (and our system's performance) and the data acquisition time.
    Our experiments explore how the performance degrades when only a small set of close views with limited view coverage of the scene are considered (mid) or a sparse set of randomly sampled views (right). }
    \label{fig:camera_views}
\end{figure}

While the quality of the final trained Instant-NGP is dependent on many factors, one of the key elements is the number of views and how well they can characterise the scene (see \reffig{fig:df_degrad}).
For applications targeting extremely high fidelity, a significant amount of effort can be dedicated to collecting many views of the scene with good coverage so that Instant-NGP performs the best.
However, downstream tasks employing a robotic perception system are often time-bounded and thus only a limited number of views are expected available at any time.
Here we explore how our system is impacted when a dense and lengthy scanning of the scene is not possible (see \reffig{fig:camera_views}).
Given that our method is limited in its capability to initialise pose hypothesis on degraded density fields (see \reffig{fig:failure_case}), in the following results we initialise all the objects using the original set of densely captured views although only a limited subset of them is used in the training of Instant-NGP used later during the pose refinement optimisation.
As per the quantitative experiment from \refsec{sec:experiments:accuracy} we report the same median translational and rotational error in the scene for which we have ground-truth object poses. 

\begin{figure}[h]
  \includegraphics[width=\linewidth]{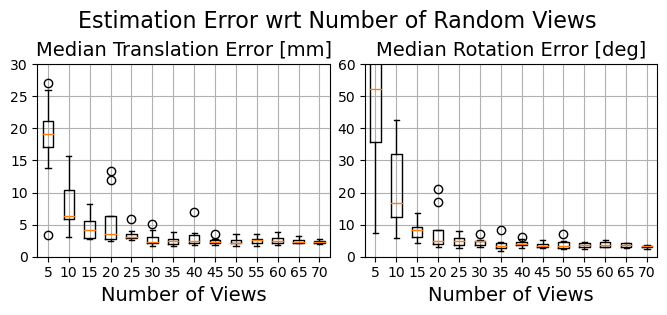}  
  \vspace{0.15cm}\includegraphics[width=\linewidth]{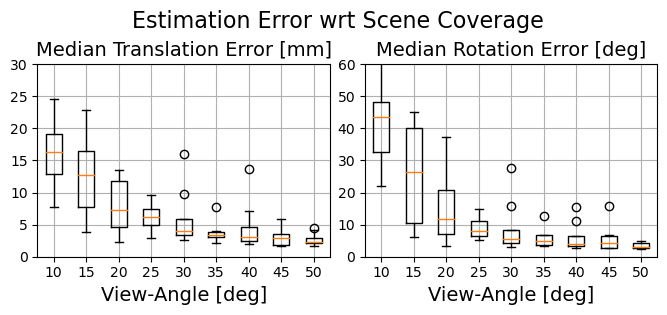}  
  % \vspace{0.15cm}\includegraphics[width=\linewidth]{images/experimental_evaluation/quantitative/nerf_iterations_exp.png}  
  \caption{Impact on pose estimation accuracy by randomly selecting a subset of the training views (top) and  varying degrees of scene coverage (bottom).
  Results here aggregate the information of 8 randomly executed experiments under the same configuration.
  }
  \label{fig:quantitative}
\end{figure}

\textbf{\textit{Sparse Random Views.}} 
Here we randomly subsample the set of all the originally captured views that densely observe the the scene.
For each of these subsets, we have a degraded yet generally good sparse coverage of the scene.
\reffig{fig:quantitative} (top) illustrates how a minimal set of views is required to accurately reconstruct the scene to any meaningful degree so that our algorithm can be applied (around 20 views). 
Beyond the minimal set, only diminishing returns are achieved to a consistent transnational and rotational error of less than 5 mm and 5 degrees, respectively. 
%This could help to guide the number of views needed to capture from a scene, speeding up scanning.

% \textit{Number of training views} impact the quality of the reconstructed Nerf, as more supervision data is available the better novel synthesis can be and thus the density field, which is of our concern in this case. Many works CITE have been tackling the problem of having good reconstructions with a limited number of views. In this experiment we train a Nerf with a different number of views. We train N Nerfs each with a different number of view. For each Nerf, we prune the original set of camera frames by dropping $N=\{1, 2, .., 15\}$ frames between consecutive frames as seen in \ref{fig:camera_views}.

\textbf{\textit{Views with Limited Scene Coverage.}}
It is not unusual that robots are not fully capable of observing the scene from an extensive number of viewpoints, for instance, due to limited reach or physical obstacles.
Here we explore how the system is impacted by these situations by first randomly selecting a viewpoint looking at the center of the scene.
Then only a subset of all the poses that are within a specified angle from the reference viewpoint are considered to train Instant-NGP.
In this experiment, the scene coverage is poorer than using a random set of sparse views for the same number of training images, but the perceived parts of the scene are to be better reconstructed.
However, this setup also imitates better the expected challenges encountered by a  robotic perception system deployed in the wild.
The presented results in \reffig{fig:quantitative} (bottom) indicate a strong dependency on this scene coverage factor, which is to be expected due to the fundamental limitations of multi-view geometry estimation.

% \textit{The number of optimisation steps} of Instant-NGP also affects reconstruction quality. However, for practical robotic applications, time and compute resources are constrained, thus potentially limiting convergence. In this experiment, we evaluate the pose estimation accuracy with Instant-NGP trained with different numbers of optimisation steps.

% We show in Figure \ref{fig:quantitative}, that pose estimation error drops the more the Nerf is trained for. This is a result of having a more well-defined and accurate density field making, allowing our fitness function to find a more accurate pose. When the density field is fuzzy early in the optimisation steps it provides a better convergence basin for our fitting function, however, this yields a lower accurate estimate due to the density field being less defined. We also show in figure \ref{fig:df_degrad} examples of the density field at various optimisation steps.

\begin{comment}
To justify our decision of running all the evaluations with the same initialisation we also provide a variant of this experiment where we initialise object poses based off the depth map available from the Nerf being fitted to. We show the results in INSERT FIG.
\end{comment}

\begin{comment}
    
\begin{figure}[h]
  \includegraphics[width=\linewidth]{images/dropping_veiws.png}
  \caption{Effect of using different number of training frames on the pose estimation accuracy. For each data point, we randomly select $n$ frames from the full set of frames available. Each number of frames evaluation, is run 8 times with different $n$ random frames.}
  \label{fig:n_frames}
\end{figure}

\begin{figure}[h]
  \includegraphics[width=\linewidth]{images/reducing_baseline.png}
  \caption{Effect of training Intsant-NGP with different number of optimisation steps on the pose estimation accuracy.}
  \label{fig:opt_steps}
\end{figure}

\begin{figure}[h]
  \includegraphics[width=\linewidth]{images/dropping_veiws.png}
  \caption{Effect of training Instant-NGP with different views with constrained field of view of the scene. We randomly select a frame in the scene and use all the frames that fall within an angular range from the the view point.}
  \label{fig:field_of_view}
\end{figure}
\end{comment}

\subsection{Ablation Study}
\label{sec:experiment:ablation}
In this section, we provide an ablation over multiple design choices, reporting in \reftab{table:ablations} the resulting estimation error in the dataset with available ground truth.

We show that the camera pose obtained from the robot arm is high to a degree that renders the density field unusable for fitting objects using our method, producing estimates that more than the object diameters.
% Although our view poses of the captured images are assumed to be retrievable directly using the robot arm's forward kinematic chain the hand-eye calibration, in practice, they are still significantly noisy for a high-fidelity reconstruction of the scene, as reported when these initial poses are not further optimised during the training of Instant-NGP.

The optimisation function \refeq{eq:alignment} requires both points on the objects' model surface  $\pointsurfaceset$ and along the normals $\pointnormalset$ to lie in high-density and low-density regions of the density field.
While it would be expected that the points on the model's surface would be sufficient to accurately estimate object pose, in practice, this is evidenced as insufficient.
We believe that using both points on the model surface $\pointsurfaceset$ and along their normal $\pointnormalset$ alleviates the fact that the resulting Instant-NGP cannot accurately model the object's geometry but can roughly estimate the transition between high-density and low-density regions.

% Optimising our proposed fitness function without samples along the normal and just naively optimising for maximising the the density for the points on the surface leads to objects getting moved into unobserved regions where the density field is noisy and where the global maxima in this case is not necessarily the pose that yields the best alignment of the object model to the reconstructed density field.

\begin{table}[t]
\centering
\begin{tabular}{lcc}
\hline
                & Trans. Error [mm] & Rot. Error [deg] \\ \hline
w/o camera pose optimisation & 17.3 & 49.4 \\ 
w/o samples along the normal & 43.2 & 70.1 \\ 
All                          &  \textbf{1.6} & \textbf{3.3} \\ \hline
\end{tabular}
\caption{Performance for different variants of the pipeline.}
\label{table:ablations}
\end{table}

\section{Conclusions}

We have presented a complete system with accurate and robust performance for model-based object pose estimation, suitable for operation in a high-precision manipulation setting where a robot arm is equipped with a single RGB camera. 
We make use of a state-of-the-art light field reconstruction method integrated with and calibrated against arm kinematics. 
Fit-NGP can estimate the full 3D pose of even small, metallic objects such as bolts and washers to within millimetre precision.
Future work would expand on improving the performance of the system and, in particular, the initialisation of the pose hypotheses when considering only a small number of posed images for which we plan to explore active and data-driven approaches to scene scanning.

% or the possibility of obtaining object's poses on-the-fly during the scanning of the scene.

% A clear weakness of our system is the need to make multi-view scans of a scene before performing pose estimation, but our ablation results indicate that good performance can often still be achieved with fewer images if they are carefully chosen. In the near future we plan to implement camera trajectories optimised to obtain the best reconstructions for pose estimation, and even to investigate active scanning, where the robot could return to capture more images if pose estimation is proving difficult.

%\begin{itemize}
%\item Extensions with other losses: RGB /Silhouette / Depth / Normals supervision loss
%\item Tightly coupled optimization. Model fitting  to also improve nerf+poses convergence / accuracy 
%\end{itemize}

\addtolength{\textheight}{-12cm}   % This command serves to balance the column lengths
                                  % on the last page of the document manually. It shortens
                                  % the textheight of the last page by a suitable amount.
                                  % This command does not take effect until the next page
                                  % so it should come on the page before the last. Make
                                  % sure that you do not shorten the textheight too much.

%%%%%%%%%%%%%%%%%%%%%%%%%%%%%%%%%%%%%%%%%%%%%%%%%%%%%%%%%%%%%%%%%%%%%%%%%%%%%%%%

%%%%%%%%%%%%%%%%%%%%%%%%%%%%%%%%%%%%%%%%%%%%%%%%%%%%%%%%%%%%%%%%%%%%%%%%%%%%%%%%

%%%%%%%%%%%%%%%%%%%%%%%%%%%%%%%%%%%%%%%%%%%%%%%%%%%%%%%%%%%%%%%%%%%%%%%%%%%%%%%%
%\section*{APPENDIX}

%Appendixes should appear before the acknowledgment.

\section{Acknowledgements}
Research presented here has been supported by Dyson Technology Ltd.

%%%%%%%%%%%%%%%%%%%%%%%%%%%%%%%%%%%%%%%%%%%%%%%%%%%%%%%%%%%%%%%%%%%%%%%%%%%%%%%%

\bibliographystyle{plain} % We choose the "plain" reference style
\bibliography{robotvision, fitngp}

\end{document}